\pdfoutput=1

\documentclass[11pt]{article}
\usepackage{tabularx}
\usepackage{booktabs}
\usepackage{multirow}
\usepackage{multicol}
\usepackage[preprint]{acl} 

\usepackage{times}
\usepackage{latexsym}

\usepackage[T1]{fontenc}

\usepackage[utf8]{inputenc}

\usepackage{microtype}

\usepackage{inconsolata}

\usepackage{graphicx}

\usepackage{enumitem}
\usepackage{amsmath}
\usepackage{calc}

\usepackage{numprint}

\npdecimalsign{.}
\npthousandsep{,}

\usepackage{makecell}

\usepackage{siunitx}
\usepackage{listings}
\usepackage{stfloats}

\title{Rethinking the Role of Text Complexity in Language Model Pretraining}

\author{
    Dan John Velasco$^{\ast}$ \and Matthew Theodore Roque$^{\ast}$ \\
    Samsung R\&D Institute Philippines\\
  \texttt{\{dj.velasco,roque.mt\}@samsung.com} \\
  \small{\textbf{$^{\ast}$Equal Contribution}} \\
  }

\begin{document}
\maketitle
\begin{abstract}
Improving pretraining data quality and size is known to boost downstream performance, but the role of text complexity—how hard a text is to read—remains less explored. We reduce surface-level complexity (shorter sentences, simpler words, simpler structure) while keeping core content approximately constant and ask: (i) How does complexity affect language modeling across model sizes? (ii) Can useful representations be learned from simpler text alone? (iii) How does pretraining text complexity influence downstream language understanding? We simplify human-written texts using a large language model, pretrain causal models (28M–500M) from scratch on original vs.\ simplified data, and evaluate them in fine-tuning and zero-shot setups. We find that perplexity is sensitive to the interaction between model capacity and text complexity—smaller models degrade far less on simpler texts—while text complexity has little impact on fine-tuning evaluations, with zero-shot evaluations indicating that simpler texts benefit performance on linguistic knowledge tasks, whereas more complex texts favor tasks requiring world knowledge and entity tracking. Our findings suggest that different types of data diversity affect transfer and zero-shot performance differently, providing insight into tailoring data curation to specific goals.
\end{abstract}

\begin{figure}[t]
    \centering
    \includegraphics[width=\columnwidth]{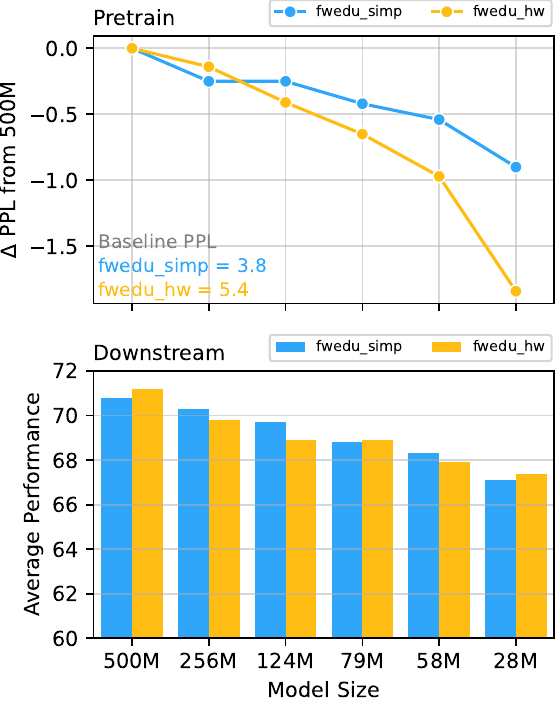}
    \caption{\textbf{(Top)} Perplexity (PPL) degrades faster for models trained on \texttt{fwedu\_hw} (human-written) than on \texttt{fwedu\_simp} (simplified) as model size decreases, suggesting that smaller models handle lower-complexity text more effectively. \textbf{(Bottom)} Average performance across 7 language tasks remains similar across data setups suggesting text complexity has limited impact on general language understanding.}
    \label{fig:main-results}
\end{figure}

\section{Introduction}

Let's compare two versions of text: 

\begin{description}[leftmargin=!,labelwidth=\widthof{(B)}]
    \item[(A)] As the sunset cast its warm orange glow over Manila Bay, people relaxed on the sideline benches, enjoying the peaceful view of the sunset.
    \item[(B)] The sunset gave Manila Bay a warm, orange light. People sat on the benches and enjoyed the view of the sunset.
\end{description}

The two versions convey the same core meaning, but one uses more nuanced, complex language, whereas the other is simpler and less nuanced. This can be likened to lossy compression, where version (B) requires fewer bits to represent the information in (A) but loses some nuance. It compresses by using common words and simpler sentence structures while retaining the core information.

\pagebreak
What if our corpus is more like (B)? Can we still learn useful representations by training solely on simplified text with a simpler vocabulary and sentence structure? To answer this, we manipulate surface-level complexity—shorter sentences, simpler words, simpler structures—while keeping the semantics close to constant, and measure downstream performance.

It is well-known that language models acquire world knowledge during pretraining \cite{petroni-etal-2019-language,roberts-etal-2020-much,zhang-etal-2021-need,wei2022emergentabilitieslargelanguage}, and transfer learning is more effective when the pretraining corpus aligns with the target task domain \cite{ruder-plank-2017-learning,gururangan-etal-2020-dont}. For example, pretraining on medical-related texts leads to better performance on medical domain tasks than using finance-related texts. This highlights that a model's knowledge base strongly affects downstream results. To isolate the effect of text complexity, it's essential to control for core content. In this paper, we ask three core questions:

\begin{enumerate}
  \item[(1)] How does text complexity affect language modeling performance across models of varying capacity?
  \item[(2)] Can we learn useful representations by training solely on simpler text, with simpler vocabulary and sentence structure?
  \item[(3)] How does the text complexity of pretraining data affect downstream performance on language understanding tasks?
\end{enumerate}

We collected human-written texts and used a Large Language Model (LLM) to produce simplified versions while preserving core content. Causal language models (28M-500M) were then pretrained from scratch in two setups: on the original texts and on their simplified counterparts. We evaluated language understanding through fine-tuning, and linguistic knowledge and commonsense reasoning in zero-shot settings.

Our empirical evidence shows that reducing surface-level complexity features does not significantly impact performance on general language understanding tasks (Figure \ref{fig:main-results}). These results suggest that text complexity is not the primary driver of performance; instead, knowledge coverage may matter more. However, zero-shot evaluations (Table \ref{tab:ling-benchmarks} and \ref{tab:reasoning-benchmarks}) suggest that text simplicity can boost performance in linguistic knowledge tasks, while greater complexity tends to aid world knowledge and entity tracking.


\section{Related Work}
\textbf{Text complexity or readability.} It refers to how difficult a text is to understand \cite{dubay2004principles}, influenced by linguistic factors such as word choice (e.g., "utilize" vs. "use"), sentence structure (complex vs. simple), and content type (academic vs. children's books) \cite{dale1948formula,graesser2004coh}. Other factors such as the reader's knowledge affect readability \cite{ozuru2009prior}. In this work, we focus solely on linguistic aspects.

Common readability metrics—such as Flesch Reading Ease (FRE) \cite{flesch1948new}, Dale-Chall \cite{dale1948formula}, and SMOG \cite{mc1969smog}—use surface features like sentence length and word complexity. These measures overlook deeper dimensions such as coherence and style, motivating machine learning and deep learning approaches \cite{hancke2012readability,meng2020readnet,imperial2021bert,chatzipanagiotidis2021broad}. Recent work applies LLMs to readability estimation \cite{trott-riviere-2024-measuring,lee-lee-2023-prompt,rooein-etal-2024-beyond}, achieving strong alignment with human judgments even without fine-tuning. However, LLM-based scoring is computationally costly at corpus scale, so we use FRE to estimate readability.

\paragraph{Text simplification (TS).} It aims to make text easier to understand while preserving its content \cite{agrawal-carpuat-2023-controlling,alva-manchego-etal-2019-cross,truicua2023simplex}. While simplified texts tend to be shorter, this is not always the case \cite{shardlow2014survey}. This is different from Text Summarization, where the goal is to shorten the text even if it changes the organization and content. Early approaches used word substitution with lexicons \cite{saggion2017automatic,shardlow2014survey,kriz2018simplification}, while others framed TS as statistical machine translation (SMT) \cite{wubben2012sentence,scarton2018text,specia2010translating,xu2016optimizing}. Subsequent work applied deep learning encoder-decoder models \cite{zhang-lapata-2017-sentence,alva-manchego-etal-2019-cross,agrawal-carpuat-2023-controlling}, and recent studies explore LLMs \cite{trott-riviere-2024-measuring,imperial-tayyar-madabushi-2023-flesch,farajidizaji-etal-2024-possible,padovani-etal-2024-automatic}. While some research targets specific grade levels, we follow \citet{trott-riviere-2024-measuring} in simplifying complex texts without grade constraints, leveraging LLMs for this task.

\paragraph{Pretraining language models on simple texts.} Recent work has explored pretraining small language models (SLMs) on simple texts. \citet{huebner-etal-2021-babyberta} showed that models trained on child-directed speech match larger models on probing tasks. \citet{eldan2023tinystoriessmalllanguagemodels} found that SLMs trained on synthetic short stories using only words familiar to 3-4-year-olds can generate coherent, fluent text. Other studies \cite{deshpande-etal-2023-honey,muckatira-etal-2024-emergent} reported that SLMs pretrained on simplified language perform comparably to larger models when problems are reformulated in simpler terms. The BabyLM Challenge \cite{warstadt-etal-2023-findings,hu2024findingssecondbabylmchallenge} pretrains language models on <100M words from child-directed and simplified texts, using provided or custom datasets within the budget.

\paragraph{Pretraining dataset design.} Large-scale pretraining is a key driver of modern language model performance \cite{NEURIPS2020_1457c0d6,kaplan2020scalinglawsneurallanguage,hoffmann2022trainingcomputeoptimallargelanguage}. Dataset design choices—domain composition, quality and toxicity filtering, and collection date—affect performance in ways that fine-tuning cannot fully correct \cite{longpre-etal-2024-pretrainers}.

The work most related to ours, \citet{agrawal-singh-2023-corpus}, finds that models trained on more complex text (e.g., Wikipedia) outperform those trained on simpler text (e.g., children's books), with complexity measured by Flesch Reading Ease. However, because they compare entirely different corpora, complexity is confounded with other factors such as topic breadth, register, discourse structure, and domain diversity. We instead manipulate complexity within the same source texts, preserving core content while varying only surface-level features. This controlled setup isolates the effect of textual complexity, complementing the broader correlation observed by \citet{agrawal-singh-2023-corpus}.

Although prior work reports positive results for simple-text pretraining, no study has directly examined the impact of surface-level complexity at larger scales (e.g., 2B tokens). Our experiments address this gap, providing empirical evidence on whether useful models can be trained solely on simplified text.

\section{Creating the Pretraining Datasets}
\subsection{Human-Written Corpus}
We took a subset of FineWeb-Edu \cite{penedo2024finewebdatasetsdecantingweb}, a collection of high-quality English web pages specifically optimized for educational content. This dataset is known for its permissive license\footnote{ODC-By 1.0} and its quality since it has gone through rigorous processing such as filtering, deduplication, and curation. The subset has 2 billion tokens\footnote{Token counts derived from Llama 2 tokenizer \cite{touvron2023llama2openfoundation}}, denoted as \texttt{\textbf{fwedu\_hw}} (short for FineWeb-Edu human-written). The choice of dataset size is motivated by Chinchilla Compute-Optimal guideline of 1:20 parameter-tokens ratio \cite{hoffmann2022trainingcomputeoptimallargelanguage} and practical reasons (e.g. training under fixed compute budget). While this size is not exactly compliant to the Chinchilla guideline for the 124M, 256M and 500M models, it is at least Chinchilla Optimal for the smaller models (e.g. 28M, 58M, 79M).

\subsection{Simplified Corpus}
We prompt Llama 3.1 8B \cite{grattafiori2024llama3herdmodels} to transform \texttt{fwedu\_hw} into simplified texts. For efficient inference, we use the INT8 quantized version\footnote{ Model accessed at:\\\url{https://huggingface.co/neuralmagic/Meta-Llama-3.1-8B-Instruct-quantized.w8a8}} of the model and vLLM \cite{kwon2023efficientmemorymanagementlarge} as our LLM serving system. We discuss more about the prompt engineering and include the final prompt in Appendix \ref{sec:prompt}.

We split the documents from \texttt{fwedu\_hw} into paragraphs\footnote{We use the term "paragraph" to refer to the smallest unit of block of text in our data pipeline. It is \textbf{not} always the case that the smallest unit is an actual paragraph. It can be a single sentence, table, heading, author lists, or other text artifacts.}. Transformation is done at the \textbf{paragraph level} because the model tends to summarize rather than simplify if the input is a multi-paragraph document. However, not all paragraphs are transformed. This can happen under three conditions: (1) when a paragraph is too short relative to its full document (e.g. title, headers); (2) when a paragraph is too long (e.g. tables, lists); or (3) when the transformation is significantly shorter or longer than the original text (e.g. incorrect simplification). In all of these cases, we \textbf{removed these paragraphs from both datasets}. This allows for more control on text complexity of the datasets while controlling for text content by keeping both datasets perfectly parallel. On both datasets, the paragraphs were not reconstructed back to document-level and instead, pretraining is done at the paragraph-level. We include a more detailed breakdown of these conditions in Appendix \ref{sec:skip_transform}.

The final simplified corpus, denoted as \texttt{\textbf{fwedu\_simp}} (short for FineWeb-Edu simplified), has around 1.71B tokens. To get a rough idea of what the simplified texts look like, see the following example:

\begin{quote}
    \textbf{Original}: Your comment really helped me feel better the most. I was sitting in my office, feeling so bad that I didn't say how inappropriate and out of line his comments were, and this helped.

    \textbf{Simplified}: Your comment really helped me feel better. I was feeling bad because I didn't speak up when someone made inappropriate comments.
\end{quote}

\section{Experimental Setup}
In our study, we investigate the effect of text complexity on pretraining and downstream performance of language models across varying model capacity. We compare models trained on \texttt{fwedu\_hw} (human-written) with those trained on \texttt{fwedu\_simp} (simplified).

\subsection{Model Architecture}
Our model architecture is based on the design choices of MobileLLM \cite{liu2024mobilellmoptimizingsubbillionparameter}: deep-and-thin architectures, SwiGLU activation \cite{shazeer2020gluvariantsimprovetransformer}, grouped-query attention \cite{ainslie-etal-2023-gqa}, and embeddings sharing \cite{press-wolf-2017-using}. All models share the same set of architectural details and hyperparameters as MobileLLM except where explicitly varied. We removed embeddings sharing for the sole purpose of making the results more generalizable to most contemporary causal language models which do not use embeddings sharing. Architectural details are summarized in Table \ref{tab:model-config}.

\subsection{Pretraining Configurations}
\begin{table}[t]
\small
\centering
\setlength{\tabcolsep}{3pt}  
\begin{tabularx}{\columnwidth}{cccccX}
\toprule
\makecell{\textbf{\#Params} \\ \textbf{(Non-Emb)}} & \textbf{\#Layer} & \textbf{\#Head} & \textbf{\#KV} & \textbf{Emb Dim} & \textbf{\#Params} \\
\midrule
500M & 40 & 18 & 6 & 1044 & \textasciitilde531M \\
256M & 30 & 9 & 3 & 846  & \textasciitilde283M \\
124M & 30 & 9 & 3 & 576  & \textasciitilde143M \\
79M  & 30 & 9 & 3 & 450  & \textasciitilde94M  \\
58M  & 30 & 9 & 3 & 378  & \textasciitilde70M  \\
28M  & 30 & 9 & 3 & 252  & \textasciitilde36M  \\
\bottomrule
\end{tabularx}
\caption{Model architecture configurations. Emb Dim is the embedding size, Non-Emb refers to non-embedding parameters, and \#KV denotes key-value heads. The number of layers and attention heads is fixed for models up to 256M and increased for the 500M model. This design maintains a consistent deep-and-thin architecture while scaling parameter count.}
\label{tab:model-config}
\end{table}

All models are trained for one epoch on either \texttt{fwedu\_hw} or \texttt{fwedu\_simp}. Both use the LLaMA-2 BPE tokenizer \citep{touvron2023llama2openfoundation} with a 32k vocabulary. Training examples are individual paragraphs, with no concatenation or sequence packing.

Inputs are right-padded to 512 tokens with the EOS token (identical to BOS) and use a causal attention mask. Each corpus is trained on independently and remains perfectly parallel after filtering, ensuring differences in model behavior stem solely from surface-level complexity.

Optimization uses AdamW \cite{loshchilov2019decoupledweightdecayregularization} with default hyperparameters, a peak learning rate of $3\mathrm{e}{-4}$ (28M models) or $5\mathrm{e}{-4}$ (all other models), linearly decayed, 5\% warm-up, and no dropout. Models with 28M-124M parameters use an effective batch size of 256 (8 examples/GPU × 8 GPUs × 4 gradient accumulation steps). Models with 256M-500M parameters use 4 examples/GPU with 8 accumulation steps to match the batch size. Training is performed in FP16 mixed precision on 8× NVIDIA P100 GPUs; gradient checkpointing is enabled only for the 500M model.

Validation is run on a held-out corpus slice after 300M tokens and at every subsequent doubling. Results are from the final checkpoint. Implementation uses PyTorch and Hugging Face Transformers. All runs fix the random seed to 42 for data shuffling and initialization.

\subsection{Fine-tuning Tasks}
To assess the downstream impact of text complexity during pretraining, we fine-tune our models on a suite of seven language understanding tasks drawn from GLUE and SuperGLUE: BoolQ, MNLI, MRPC, MultiRC, QQP, RTE, and WSC. This set follows the evaluation configuration of the BabyLM Challenge, and we use the same preprocessed datasets provided in the BabyLM evaluation pipeline for both training and validation \cite{charpentier2025babylmturns3papers}.

All models are trained with an added classification head. For tasks involving multiple input sequences (e.g., premise-hypothesis or sentence pairs), we concatenate the two sequences with a separator token before feeding them into the model. We perform two fine-tuning regimes for each model-task pair:
(1) full-model fine-tuning, where all model parameters and the classification head are updated; and
(2) linear probing, where only the classification head is updated.

Fine-tuning uses 8×P100 GPUs without gradient accumulation. Batch size is determined by GPU memory constraints: for BoolQ and MultiRC we use 2 examples per GPU (effective batch size of 16), and for all other tasks we use 8 examples per GPU (effective batch size of 64). For each task, we perform a grid search over learning rates ${1\mathrm{e}{-4}, 5\mathrm{e}{-5}, 2\mathrm{e}{-5}, 1\mathrm{e}{-5}, 5\mathrm{e}{-6}}$ and training epochs ${1, 2, 3, 4, 5}$.

We use the same task metrics as the BabyLM evaluations, namely: 3-class accuracy for MNLI; binary accuracy for BoolQ, MultiRC, and WSC; and F1-Score for MRPC and QQP. All experiments are run with three random seeds; we report the mean and standard deviation of the best-performing configuration for each seed. Fine-tuning is performed for all model sizes on all tasks under both training regimes.

\subsection{Zero-shot Tasks}
To further evaluate the quality of representations, we conduct zero-shot evaluations on eight multiple-choice benchmarks, grouped into two categories:

\textbf{Linguistic knowledge, entity tracking, and world knowledge}: BLiMP \citep{warstadt2020blimp} and the BLiMP Supplement (as provided in the BabyLM evaluation pipeline), probe syntactic and morphological phenomena. Entity Tracking \citep{kim-schuster-2023-entity} and EWoK (Elements of World Knowledge \citealp{ivanova2024elements}), measure a model's ability to follow discourse entities and recall factual knowledge. The models were evaluated on these tasks using the BabyLM evaluation pipeline \cite{charpentier2025babylmturns3papers}.

\textbf{Commonsense reasoning}: ARC-Easy and ARC-Challenge \citep{allenai:arc}, HellaSwag \citep{zellers2019hellaswag}, and PIQA \citep{Bisk2020}, require reasoning over everyday scenarios and physical commonsense. The models were evaluated on these tasks using the \texttt{lm-evaluation-harness} \cite{eval-harness}.

All tasks are multiple-choice. For each evaluation instance, we format the input according to the benchmark's specifications and score each candidate option by the sum of log-probabilities of its tokens given the prompt. The option with the highest score is selected as the model's prediction. 

We report accuracy for all zero-shot tasks. Evaluation is deterministic, as predictions depend solely on model likelihoods and not on sampling.

\begin{figure*}[t]
  \centering
  \includegraphics[width=\textwidth]{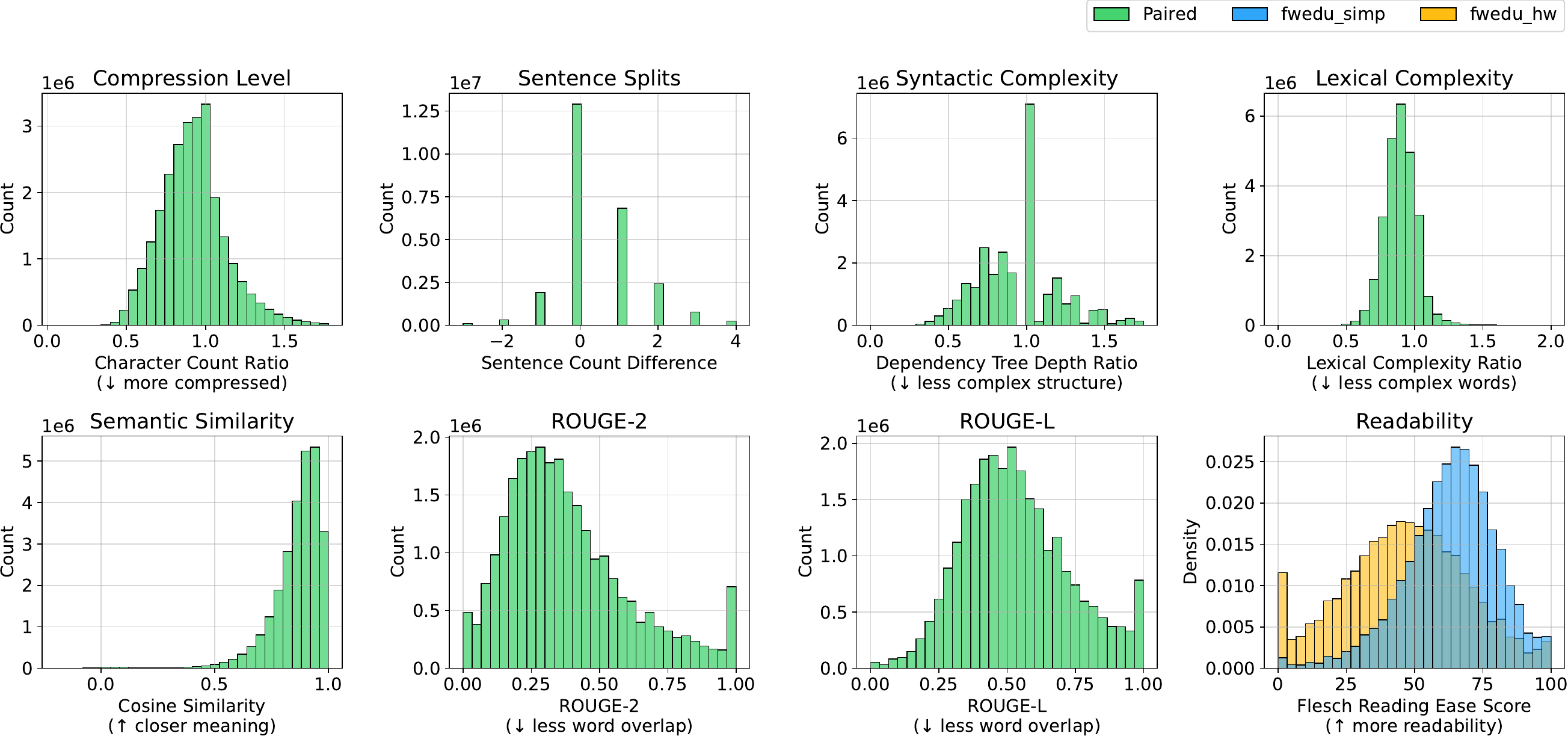}
  \caption{Corpus Features distribution. First row shows metrics of \texttt{fwedu\_simp} to \texttt{fwedu\_hw}. Second row are pairwise metrics except for Flesch Reading Ease (FRE) which only requires one input. The first row suggests \texttt{fwedu\_simp} is shorter, has more sentences, uses simpler structures, and more common words. The second row shows that \texttt{fwedu\_simp} is semantically similar to \texttt{fwedu\_hw}, with low word-order overlap (low ROUGE-2), moderate preservation of idea flow and structure (moderate ROUGE-L), and clearly higher FRE, indicating systematic differences in readability. For visualization, we removed outliers, which account for only 2.9\% of the data (see Appendix \ref{sec:ratio-outliers} for definition and examples of outliers).}
  \label{fig:corpus-metrics-plots}
\end{figure*}

\section{Results}
We performed three independent runs with different random seeds. For each run, we selected the best result over our fixed hyperparameter grid, and report the average of those three best scores. Random seeds were fixed for full reproducibility.

\subsection{Dataset Complexity Verification}
Is the simplified corpus truly simpler? To answer, we compute per-dataset and cross-dataset metrics (Table~\ref{tab:corpus-metrics-table}) and analyzed their distributions (Figure~\ref{fig:corpus-metrics-plots}). The simplified corpus has fewer tokens, a smaller vocabulary (Types), lower lexical diversity (Type-Token Ratio), and reduced unpredictability (Unigram Entropy)—all indicating lower text complexity. Cross-dataset metrics show that 26.62\% of the data are more concise, ~92\% exhibits low to medium lexical overlap (ROUGE-2), and 79\% retains at least 80\% semantic similarity (Cosine Similarity). These results suggest that the simplified dataset \textbf{differs in form while preserving core content}.

\begin{table}[ht]
\footnotesize
\centering
\begin{tabularx}{\columnwidth}{@{}l S[table-format=2.2] S[table-format=2.2]@{}}

\toprule
\textbf{Feature} & \textbf{Simplified} & \textbf{Human-written} \\
\midrule

\multicolumn{3}{@{}l}{\textbf{\textsc{Per-dataset Stats}}} \\
Total tokens & 1.71B & 2.00B \\
Total words & 1.44B & 1.57B \\
Types (unique words) & 2.76M & 5.23M \\
Type-token ratio (\%)  & 0.19\% & 0.33\% \\
Unigram entropy (bits) & 9.87 & 10.58 \\

\midrule
\multicolumn{3}{@{}l}{\textbf{\textsc{Cross-dataset Stats}}} \\
Compression ($<$80\%) & 26.62\% & \text{\textemdash} \\
Exact match & 2.51\% & \text{\textemdash} \\
High lexical overlap & 6.47\% & \text{\textemdash} \\
Medium lexical overlap & 31.13\% & \text{\textemdash} \\
Low lexical overlap & 61.75\% & \text{\textemdash} \\
Exact mismatch & 1.56\% & \text{\textemdash} \\
Semantic Sim ($>$80\%) & 79.00\% & \text{\textemdash} \\
\bottomrule
\end{tabularx}
\caption{Per-dataset and Cross-dataset statistics. Reduced per-dataset stats in Simplified indicate lower complexity compared to Human-written. Lexical overlap is measured using ROUGE-2 (R2), with the following thresholds: exact match ($R2 = 1$), high ($0.8 < R2 < 1$), medium ($0.4 < R2 \leq 0.8$), low ($0 < R2 \leq 0.4$), and exact mismatch ($R2 = 0$). Semantic Sim is computed as cosine similarity of paragraph embeddings. Cross-dataset stats suggest Simplified texts differ in form but preserve core content.}
\label{tab:corpus-metrics-table}
\end{table}

Figure \ref{fig:corpus-metrics-plots} compares the distributions of paired metrics for \texttt{fwedu\_simp} and \texttt{fwedu\_hw}, labeled as “Paired”. The first-row metrics are adapted from ASSET's text complexity evaluation \cite{alva-manchego-etal-2020-asset}:
\begin{itemize}
  \item \textbf{Compression Level}: ratio of character counts; values $< 1.0$ indicate more concise texts.
  \item \textbf{Sentence Splits}: difference in sentence counts; values $> 0$ indicate splitting of complex sentences into simpler ones.
  \item \textbf{Syntactic Complexity}: ratio of maximum dependency-tree depth; values $< 1.0$ indicate shallower (simpler) sentence structures.
  \item \textbf{Lexical Complexity}: mean squared log-rank of non-stopword tokens, based on the top 50k words in FastText embeddings\footnote{2 million word vectors trained on Common Crawl (600B tokens), \url{https://fasttext.cc/docs/en/english-vectors.html}} \cite{mikolov2018advances}; values $< 1.0$ indicate use of more frequent words.
\end{itemize}

\begin{table*}[t]
\centering
\setlength{\tabcolsep}{6pt}       
\renewcommand{\arraystretch}{0.9} 
\footnotesize
\begin{tabularx}{\textwidth}{@{}X*{8}{l}@{}}
\toprule
\textbf{Model} & 
\multicolumn{1}{l}{\textbf{boolq}} & 
\multicolumn{1}{l}{\textbf{mnli}} & 
\multicolumn{1}{l}{\textbf{mrpc}} & 
\multicolumn{1}{l}{\textbf{multirc}} & 
\multicolumn{1}{l}{\textbf{qqp}} & 
\multicolumn{1}{l}{\textbf{rte}} & 
\multicolumn{1}{l}{\textbf{wsc}} & 
\multicolumn{1}{l}{\textbf{Avg.}} \\
\midrule
\textbf{Majority Baseline} & 64.0 & 33.1 & 68.1 & 57.5 & 62.7 & 53.9 & 61.5 & 57.3\\
\midrule

\textbf{28M} & & & & & & & & \\
\hspace{1em}from\_scratch & 66.9 ± 0.8 & 36.0 ± 0.5 & 70.4 ± 1.0 & 59.0 ± 0.3 & 71.3 ± 0.3 & 58.3 ± 2.1 & 65.5 ± 2.1 & 61.1 \\
\hspace{1em}fwedu\_hw & 69.9 ± 0.2 & 61.9 ± 0.7 & 79.0 ± 1.7 & 59.1 ± 0.2 & 78.6 ± 0.3 & 60.6 ± 2.6 & 62.5 ± 4.7 & 67.4 \\
\hspace{1em}fwedu\_simp & 69.3 ± 0.6 & 61.8 ± 0.2 & 78.2 ± 1.5 & 58.8 ± 0.2 & 78.8 ± 0.2 & 60.0 ± 2.0 & 63.1 ± 2.7 & 67.1 \\
\midrule
\textbf{58M} & & & & & & & & \\
\hspace{1em}from\_scratch & 67.5 ± 0.7 & 37.7 ± 1.2 & 70.8 ± 0.7 & 58.8 ± 0.4 & 71.8 ± 0.3 & 57.4 ± 2.9 & 64.9 ± 1.0 & 61.3 \\
\hspace{1em}fwedu\_hw & 69.6 ± 0.4 & 63.2 ± 0.5 & 79.6 ± 1.5 & 58.5 ± 0.3 & 79.6 ± 0.1 & 60.2 ± 1.4 & 64.3 ± 1.8 & 67.9 \\
\hspace{1em}fwedu\_simp & 69.6 ± 0.1 & 63.6 ± 0.2 & 77.9 ± 1.0 & 58.2 ± 0.4 & 80.2 ± 0.0 & 66.4 ± 2.4 & 61.9 ± 5.5 & 68.3 \\
\midrule
\textbf{79M} & & & & & & & & \\
\hspace{1em}from\_scratch & 67.4 ± 0.8 & 38.7 ± 0.3 & 70.0 ± 0.3 & 58.3 ± 0.3 & 72.1 ± 0.4 & 56.2 ± 3.7 & 65.5 ± 1.0 & 61.2 \\
\hspace{1em}fwedu\_hw & 69.9 ± 0.3 & 64.0 ± 0.3 & 79.5 ± 1.1 & 58.7 ± 0.2 & 80.4 ± 0.4 & 66.2 ± 2.1 & 63.7 ± 4.1 & 68.9 \\
\hspace{1em}fwedu\_simp & 69.3 ± 1.0 & 65.2 ± 0.3 & 80.3 ± 1.4 & 58.2 ± 0.4 & 80.7 ± 0.1 & 67.4 ± 2.5 & 60.7 ± 0.0 & 68.8 \\
\midrule
\textbf{124M} & & & & & & & & \\
\hspace{1em}from\_scratch & 68.0 ± 0.8 & 39.5 ± 1.4 & 70.8 ± 0.7 & 58.3 ± 0.5 & 71.9 ± 0.1 & 57.4 ± 4.2 & 65.5 ± 2.1 & 61.6 \\
\hspace{1em}fwedu\_hw & 70.7 ± 0.2 & 65.9 ± 0.2 & 80.3 ± 1.4 & 59.2 ± 0.0 & 80.7 ± 0.2 & 65.5 ± 3.2 & 60.1 ± 1.0 & 68.9 \\
\hspace{1em}fwedu\_simp & 70.3 ± 0.4 & 66.9 ± 0.5 & 82.1 ± 0.7 & 58.8 ± 0.3 & 81.2 ± 0.1 & 67.6 ± 1.7 & 61.3 ± 2.7 & 69.7 \\
\midrule
\textbf{256M} & & & & & & & & \\
\hspace{1em}from\_scratch & 68.0 ± 0.7 & 40.5 ± 0.9 & 71.0 ± 1.0 & 58.5 ± 0.7 & 72.1 ± 0.2 & 57.6 ± 1.0 & 65.2 ± 1.3 & 61.8 \\
\hspace{1em}fwedu\_hw & 70.8 ± 0.4 & 67.9 ± 0.4 & 80.6 ± 2.0 & 59.1 ± 0.2 & 81.4 ± 0.2 & 66.0 ± 0.7 & 62.5 ± 1.8 & 69.8 \\
\hspace{1em}fwedu\_simp & 70.8 ± 0.7 & 67.0 ± 0.5 & 81.7 ± 1.0 & 58.6 ± 0.4 & 81.5 ± 0.1 & 70.8 ± 0.7 & 61.3 ± 2.1 & 70.3 \\
\midrule
\textbf{500M} & & & & & & & & \\
\hspace{1em}from\_scratch & 68.6 ± 0.3 & 39.6 ± 0.4 & 72.4 ± 0.3 & 58.5 ± 0.0 & 72.5 ± 0.0 & 60.6 ± 2.0 & 65.5 ± 1.0 & 62.5 \\
\hspace{1em}fwedu\_hw & 70.5 ± 0.5 & 67.6 ± 0.1 & 83.4 ± 1.0 & 58.8 ± 0.2 & 82.1 ± 0.0 & 71.5 ± 0.7 & 64.3 ± 3.1 & 71.2 \\
\hspace{1em}fwedu\_simp & 70.1 ± 0.5 & 67.4 ± 0.5 & 82.7 ± 0.7 & 58.7 ± 0.3 & 81.6 ± 0.1 & 71.8 ± 1.1 & 63.1 ± 2.1 & 70.8 \\
\bottomrule
\end{tabularx}
\caption{Full fine-tuning performance on 7 NLU tasks. Average accuracy across tasks is reported over 3 runs. The Avg. column reports mean accuracy over available tasks. Task metrics are as follows: 3-class accuracy (MNLI), Binary Accuracy (BoolQ, MultiRC, WSC), and F1-Score (MRPC, QQP). Overall results shows minimal performance difference across pretraining setups regardless of model size which suggests text complexity have minimal impact on general language understanding tasks.}
\label{tab:nlu-table}
\end{table*}

The second row of Figure \ref{fig:corpus-metrics-plots} shows semantic similarity, word overlap, and readability. We used \texttt{all-MiniLM-L6-v2} to encode paragraph embeddings, optimized for tasks like sentence similarity and clustering\footnote{\texttt{all-MiniLM-L6-v2} ranks 1st and 17th on the MTEB leaderboard \cite{muennighoff-etal-2023-mteb} for <100M and <1B parameter models.}. High scores near 1.0 indicate most simplified paragraphs retain the original meaning.

ROUGE was computed with the Evaluate tool \cite{von-werra-etal-2022-evaluate}, scoring from 0 to 1. ROUGE-2 measures exact bigram overlap, reflecting local phrasing; most examples score 0-0.4, indicating low lexical overlap. This confirms high cosine similarity reflects shared meaning, not surface form. ROUGE-L, measuring longest in-order subsequences, shows more varied scores, suggesting moderate structural similarity.

Readability was measured using Flesch Reading Ease (FRE), which factors text length, word count, and syllables. Higher FRE means simpler text: easy (60+), fairly difficult (50-60), hard (<50) \cite{readability}. \texttt{fwedu\_hw} skews lower, \texttt{fwedu\_simp} higher, indicating systematic readability differences.

Together, these results show our simplified dataset is \textbf{simpler in form while preserving core content}. For examples, see Appendix \ref{sec:examples-semantic} and \ref{sec:examples-rouge2}.

\subsection{Main Comparison: Human-Written vs. Simplified}
\subsubsection{Language-Modeling Performance}
How does text complexity affect language modeling performance across models of varying capacity? To answer this, we measured each model's perplexity degradation—defined as the absolute difference in perplexity between the 500M model and smaller models. Figure \ref{fig:main-results} shows that models trained on \texttt{fwedu\_hw} degrade faster than those trained on \texttt{fwedu\_simp} as capacity decreases, with a sharp drop for the 28M model on \texttt{fwedu\_hw}. This interaction between model capacity and data complexity suggests that future model design and selection should account for the complexity of the training data.

\subsubsection{Fine-tuning Evaluation}
The results of the full-model fine-tuning are summarized in Table \ref{tab:nlu-table}. To contextualize the impact of pretraining, we include scores from the Majority baseline and models fine-tuned from random weights (\texttt{from\_scratch}). Notably, MNLI shows the greatest gain from pretraining. More broadly, similar \texttt{from\_scratch} performance across tasks and model sizes suggests an upper bound imposed by the training data on this specific architecture—model scaling alone does not improve performance. On MultiRC, all models perform only slightly above the majority baseline, suggesting failure to learn the task. We suspect this stems from paragraph-level pretraining, which may lack signals for skills like coreference resolution which is important to succeed in MultiRC. The same likely applies to WSC.

Overall, \texttt{fwedu\_hw} and \texttt{fwedu\_simp} yield similar performance across model sizes in full-model fine-tuning. This pattern holds under linear probing as well (Table~\ref{tab:nlu-table-linearprobe} in Appendix \ref{sec:linear}), reinforcing the observation. Full-model fine-tuning also confirms the well-established trend that larger models perform better, regardless of data complexity. These results suggest that text complexity is not the primary driver of performance; instead, knowledge coverage may matter more.

\subsubsection{Zero-shot Evaluation}
\textbf{Linguistic Knowledge, Entity Tracking, and World Knowledge}. Table \ref{tab:ling-benchmarks} summarizes the zero-shot performance on linguistic knowledge and entity tracking benchmarks. BLiMP performance improves with model size, with both setups performing similarly overall, though \texttt{fwedu\_simp} has a clear edge on BLiMP-supplement. Entity tracking performance varies widely with model size; \texttt{fwedu\_hw} often leads, while \texttt{fwedu\_simp} surpasses random chance (20\%) only from 79M onward. EWoK performance improves consistently with model size, with \texttt{fwedu\_hw} often slightly outperforming \texttt{fwedu\_simp}.

\begin{table}[t]
\centering
\footnotesize
\begin{tabularx}{\columnwidth}{@{}Xcccc@{}}
\toprule
\textbf{Model} & 
\multicolumn{1}{r}{\textbf{blimp}} & 
\multicolumn{1}{r}{\textbf{blimp-supp}} & 
\multicolumn{1}{r}{\textbf{ewok}} & 
\multicolumn{1}{r}{\textbf{entity}} \\
\midrule

\textbf{28M} & & & & \\
\hspace{1em}fwedu\_hw   & 67.83 & 55.90 & 52.79 & 16.15 \\
\hspace{1em}fwedu\_simp & 66.90 & 57.47 & 52.67 & 18.78 \\
\midrule
\textbf{58M} & & & & \\
\hspace{1em}fwedu\_hw   & 69.69 & 59.03 & 52.69 & 26.35 \\
\hspace{1em}fwedu\_simp & 70.73 & 62.15 & 53.81 & 18.36 \\
\midrule
\textbf{79M} & & & & \\
\hspace{1em}fwedu\_hw   & 70.67 & 60.47 & 54.09 & 28.89 \\
\hspace{1em}fwedu\_simp & 70.44 & 61.55 & 53.09 & 20.73 \\
\midrule
\textbf{124M} & & & & \\
\hspace{1em}fwedu\_hw   & 71.64 & 62.61 & 54.07 & 25.09 \\
\hspace{1em}fwedu\_simp & 71.30 & 63.27 & 54.01 & 21.79 \\
\midrule
\textbf{256M} & & & & \\
\hspace{1em}fwedu\_hw   & 72.37 & 62.61 & 56.18 & 29.71 \\
\hspace{1em}fwedu\_simp & 72.53 & 63.65 & 55.09 & 22.58 \\
\midrule
\textbf{500M} & & & & \\
\hspace{1em}fwedu\_hw   & 72.23 & 61.85 & 56.72 & 20.81 \\
\hspace{1em}fwedu\_simp & 72.60 & 63.79 & 54.85 & 22.05 \\

\bottomrule
\end{tabularx}
\caption{
Zero-shot evaluations on grammatical knowledge (blimp), world knowledge (ewok), and Entity Tracking (entity) show consistent improvement with model size. Both setups perform similarly on BLiMP, \texttt{fwedu\_simp} scores higher on BLiMP-supplement, whereas \texttt{fwedu\_hw} leads on Entity and slightly on EWoK.
}
\label{tab:ling-benchmarks}
\end{table}

\paragraph{\textbf{Commonsense Reasoning}.} Table \ref{tab:reasoning-benchmarks} summarizes zero-shot performance on commonsense reasoning benchmarks. Performance generally improves with increased model size, especially on ARC-Easy and PIQA. ARC-Challenge remains difficult across all setups, with accuracies near random chance (20\%). This may be simply due to the pretraining data not containing the knowledge that ARC-Challenge is designed to test. On ARC-Easy, \texttt{fwedu\_hw} consistently outperforms \texttt{fwedu\_simp}, reaching a peak accuracy of 42.09\% at 256M—3 points higher than \texttt{fwedu\_simp}. On Hellaswag, both setups perform comparably across model sizes. While for PIQA, \texttt{fwedu\_simp} slightly outperforms \texttt{fwedu\_hw} consistently. Interestingly, 500M models perform worse than 256M models across tasks. We suspect this is due to the limited data size—2B tokens for \texttt{fwedu\_hw} and 1.71B for \texttt{fwedu\_simp}—bottlenecking the larger models.

\begin{table}[t]
\centering
\footnotesize
\begin{tabularx}{\columnwidth}{@{}Xcccc@{}}
\toprule
\textbf{Model} & 
\multicolumn{1}{r}{\textbf{arc\_e}} & 
\multicolumn{1}{r}{\textbf{arc\_chl}} & 
\multicolumn{1}{r}{\textbf{hellaswag}} & 
\multicolumn{1}{r}{\textbf{piqa}} \\
\midrule

\textbf{28M} & & & & \\
\hspace{1em}fwedu\_hw   & 33.33 & 20.22 & 26.99 & 56.20 \\
\hspace{1em}fwedu\_simp & 31.94 & 21.59 & 26.40 & 55.93 \\
\midrule
\textbf{58M} & & & & \\
\hspace{1em}fwedu\_hw   & 39.10 & 19.54 & 27.38 & 57.89 \\
\hspace{1em}fwedu\_simp & 33.59 & 21.42 & 27.50 & 58.22 \\
\midrule
\textbf{79M} & & & & \\
\hspace{1em}fwedu\_hw   & 38.80 & 21.25 & 27.69 & 58.71 \\
\hspace{1em}fwedu\_simp & 38.05 & 20.65 & 27.43 & 59.52 \\
\midrule
\textbf{124M} & & & & \\
\hspace{1em}fwedu\_hw   & 38.09 & 19.37 & 28.08 & 58.16 \\
\hspace{1em}fwedu\_simp & 38.85 & 21.50 & 28.36 & 60.77 \\
\midrule
\textbf{256M} & & & & \\
\hspace{1em}fwedu\_hw   & 42.09 & 22.70 & 28.85 & 60.99 \\
\hspace{1em}fwedu\_simp & 38.80 & 20.82 & 28.94 & 61.10 \\
\midrule
\textbf{500M} & & & & \\
\hspace{1em}fwedu\_hw   & 40.99 & 21.25 & 28.30 & 58.16 \\
\hspace{1em}fwedu\_simp & 33.96 & 18.43 & 27.52 & 57.99 \\

\bottomrule
\end{tabularx}
\caption{Zero-shot accuracy on commonsense reasoning benchmarks shows that ARC-Challenge (arc\_chl) remains near random chance (20\%) across all setups. All other tasks improve consistently with model size. \texttt{fwedu\_hw} performs best on ARC-Easy (arc\_e), while \texttt{fwedu\_simp} slightly outperforms on PIQA. Both setups perform similarly on HellaSwag. All 500M models show a performance drop relative to 256M across tasks.}
\label{tab:reasoning-benchmarks}
\end{table}

\section{Discussion}
In this section, we reflect on the broader implications of our findings for data curation and synthetic data generation. We frame these as conjectures rather than definitive claims.

Our experiments controlled for lexical, syntactic, semantic, and stylistic diversity, though these do not exhaust the full space of variation. Ideally, optimizing along multiple dimensions may yield broader benefits, but practical constraints often force trade-offs. Our results provide empirical evidence on which outcomes, such as transfer or zero-shot performance, are most sensitive to particular forms of diversity. This can help guide decisions when prioritizing which dimensions to optimize.  

\textbf{Data curation.} In data curation or pruning, practitioners sometimes emphasize surface-level variety (lexical or syntactic) as a proxy for diversity. Our experiments suggest this can be misleading. We find that reducing lexical and syntactic variation, while preserving topical and knowledge coverage, did not harm transfer performance but did impair zero-shot generalization. This implies that curation strategies should be designed with the intended use case in mind: fine-tuned applications may tolerate reduced surface variation, whereas zero-shot settings are more sensitive to it.  

\textbf{Synthetic data design.} Generation-based synthetic datasets often suffer from low-diversity outputs \citep{gandhi-etal-2024-better} and are vulnerable to collapse effects \citep{shumailov2024curserecursiontraininggenerated, guo2024curiousdeclinelinguisticdiversity, briesch2024largelanguagemodelssuffer}, underscoring the need for diversity-aware generation. Our findings indicate that not all forms of diversity contribute equally: different axes influence different outcomes. A practical design strategy may be to prioritize broad topic and knowledge coverage first, then deliberately introduce surface-level variety (e.g., controlled paraphrasing) to support zero-shot performance if needed.

\section{Conclusion}
In this work, we investigate how text complexity affects language model pretraining. Specifically, we ask whether simplified language—while preserving core content—can lead to representations that perform as well as those learned from more complex, human-written text. We pretrained causal language models of varying sizes (28M-500M parameters) on both simplified and human-written corpora. Our results show that simplifying surface-level features does not significantly hurt downstream performance on a range of language understanding tasks. However, models trained on more complex text show an advantage in zero-shot settings on benchmarks requiring reasoning and knowledge of the world—such as Entity Tracking, EWoK, and ARC-Easy—while performing similarly on BLiMP, HellaSwag, and PIQA. These findings highlight that different types of data diversity affect transfer and zero-shot performance differently, providing insight into tailoring data curation to specific goals.

\section*{Limitations}
Our study has several limitations. First, the LLM-based simplification process is imperfect and may introduce subtle inconsistencies in core content due to hallucinations. Second, the 2B-token corpora are relatively small by today's pretraining standards, potentially limiting model performance. Third, our fine-tuning evaluation focuses on a narrow set of classification and multiple-choice benchmarks, which may not capture the full range of model capabilities, particularly in open-ended or generative tasks. Fourth, our zero-shot evaluation may not fully reflect the targeted capabilities, as it is constrained by limited training data and model capacity. Fifth, we focus solely on causal language models, leaving open the possibility that different patterns may emerge with encoder models like BERT. Lastly, we did not conduct a per-phenomenon analysis of BLiMP, leaving open the possibility that certain linguistic constructions are more sensitive to simplification.

\bibliography{custom}

\pagebreak
\appendix

\section{Linear Probing Results}
\label{sec:linear}
\begin{table*}[!btp]
\centering
\small
\begin{tabularx}{\textwidth}{@{}X*{8}{c}@{}}
\toprule
\textbf{Model} & 
\multicolumn{1}{r}{\textbf{boolq}} & 
\multicolumn{1}{r}{\textbf{mnli}} & 
\multicolumn{1}{r}{\textbf{mrpc}} & 
\multicolumn{1}{r}{\textbf{multirc}} & 
\multicolumn{1}{r}{\textbf{qqp}} & 
\multicolumn{1}{r}{\textbf{rte}} & 
\multicolumn{1}{r}{\textbf{wsc}} & 
\multicolumn{1}{r}{\textbf{Avg.}} \\
\midrule
\textbf{Baseline} & & & & & & & & \\
\hspace{1em} majority & 64.0 & 33.1 & 68.1 & 57.5 & 62.7 & 53.9 & 61.5 & 57.3\\

\addlinespace
\textbf{28M} & & & & & & & & \\
\hspace{1em}fwedu\_hw & 65.1 ± 0.9 & 42.9 ± 0.3 & 69.9 ± 0.7 & 57.2 ± 0.6 & 69.7 ± 0.3 & 53.9 ± 2.6 & 50.6 ± 5.7 & 58.5 \\
\hspace{1em}fwedu\_simp & 65.7 ± 1.6 & 43.0 ± 2.2 & 70.7 ± 1.0 & 54.8 ± 0.5 & 69.8 ± 0.2 & 56.0 ± 3.4 & 59.5 ± 3.7 & 59.9 \\

\addlinespace
\textbf{58M} & & & & & & & & \\
\hspace{1em}fwedu\_hw & 64.4 ± 0.5 & 45.2 ± 0.5 & 69.4 ± 0.6 & 54.8 ± 0.6 & 69.2 ± 0.2 & 49.5 ± 2.9 & 62.5 ± 4.7 & 59.3 \\
\hspace{1em}fwedu\_simp & 64.2 ± 0.5 & 45.6 ± 0.3 & 69.2 ± 0.0 & 54.8 ± 0.2 & 68.1 ± 0.6 & 56.2 ± 1.8 & 56.5 ± 9.2 & 59.2 \\

\addlinespace
\textbf{79M} & & & & & & & & \\
\hspace{1em}fwedu\_hw & 66.0 ± 0.3 & 44.4 ± 1.2 & 70.7 ± 0.8 & 55.9 ± 0.3 & 69.0 ± 0.5 & 55.3 ± 3.6 & 55.4 ± 8.9 & 59.5 \\
\hspace{1em}fwedu\_simp & 64.5 ± 0.3 & 46.8 ± 0.5 & 69.7 ± 1.0 & 55.8 ± 1.6 & 68.5 ± 0.2 & 53.2 ± 1.4 & 57.1 ± 3.1 & 59.4 \\

\addlinespace
\textbf{124M} & & & & & & & & \\
\hspace{1em}fwedu\_hw & 64.5 ± 0.2 & 45.2 ± 0.4 & 70.5 ± 1.5 & 54.2 ± 0.7 & 69.9 ± 0.6 & 50.5 ± 4.0 & 57.1 ± 4.7 & 58.9 \\
\hspace{1em}fwedu\_simp & 64.5 ± 0.1 & 46.3 ± 0.2 & 70.0 ± 0.3 & 54.8 ± 0.4 & 70.3 ± 0.8 & 53.0 ± 3.3 & 57.1 ± 4.7 & 59.4 \\

\addlinespace
\textbf{256M} & & & & & & & & \\
\hspace{1em}fwedu\_hw & 65.1 ± 0.1 & 48.3 ± 0.5 & 69.7 ± 0.8 & 56.1 ± 0.9 & 68.8 ± 0.2 & 55.8 ± 1.7 & 60.7 ± 3.6 & 60.6 \\
\hspace{1em}fwedu\_simp & 65.8 ± 1.0 & 48.9 ± 0.8 & 70.0 ± 0.6 & 55.3 ± 1.3 & 70.1 ± 0.2 & 57.2 ± 2.4 & 57.1 ± 3.6 & 60.6 \\

\addlinespace
\textbf{500M} & & & & & & & & \\
\hspace{1em}fwedu\_hw & 64.9 ± 0.7 & 49.1 ± 0.7 & 70.7 ± 0.5 & 56.0 ± 0.9 & 70.1 ± 0.5 & 49.5 ± 2.4 & 63.1 ± 2.7 & 60.5 \\
\hspace{1em}fwedu\_simp & 65.0 ± 0.2 & 48.9 ± 0.4 & 71.3 ± 1.0 & 56.0 ± 1.7 & 69.2 ± 0.6 & 57.6 ± 6.7 & 59.5 ± 6.8 & 61.1 \\

\bottomrule
\end{tabularx}
\caption{Linear Probe performance on 7 NLU tasks. Average accuracy across tasks is reported over 3 runs. The Avg. column reports mean accuracy over available tasks. Overall results shows minimal performance differences between pretraining setups across model sizes. This supports the Full fine-tuning findings (Table \ref{tab:nlu-table}), suggesting that text complexity has limited impact on general language understanding tasks.}
\label{tab:nlu-table-linearprobe}
\end{table*}

\pagebreak
\section{Text Simplification Prompt}
\label{sec:prompt}

The prompt engineering is done through trial-and-error and judged by the authors according to the following qualitative criteria:

\begin{itemize}
    \item Does it use simpler words? By "simpler words," we mean commonly used words.
    \item Does it convert compound or complex sentences into simple sentences?
    \item Does it preserve the original content and organization of thoughts?
\end{itemize}

Once we found a prompt that can reliably do all those things on a small sample, we used that prompt to transform the whole corpus.

The final prompt is shown below:

\lstset{
  basicstyle=\ttfamily\small,
  breaklines=true,
  columns=fullflexible,
  frame=single,
  backgroundcolor=\color{gray!10},
}

\begin{lstlisting}
    ---
    
    Role Description:
    You are an experienced educator and linguist specializing in simplifying complex texts without losing any key information or changing the content. Your focus is to make texts more accessible and readable for primary and secondary school students, ensuring that the essential information is preserved while the language and structure are adapted for easier comprehension.
    
    ---
    
    Task Instructions:
    1. Read the Following Text Carefully:
       - Thoroughly understand the content, context, and purpose of the text to ensure all key information is retained in the simplified version.
    
    2. Simplify the Text for Primary/Secondary School Students:
       - Rewrite the text to make it more accessible and easier to understand.
       - Use age-appropriate language and simpler sentence structures.
       - Maintain all key information and do not omit any essential details.
       - Ensure that the original meaning and intent of the text remain unchanged.
    
    3. Preserve Key Information:
       - Identify all essential points, facts, and ideas in the original text.
       - Ensure these elements are clearly presented in the simplified version.
    
    4. Avoid Adding Personal Opinions or Interpretations:
       - Do not introduce new information or personal views.
       - Focus solely on simplifying the original content.
    
    ---
    
    Simplification Guidelines:
    
    Sentence Structure:
    - Use simple or compound sentences.
    - Break down long or complex sentences into shorter ones.
    - Ensure each sentence conveys a clear idea.
    
    Vocabulary:
    - Use common words familiar to primary and secondary school students.
    - Replace advanced or technical terms with simpler synonyms or provide brief explanations.
    - Avoid jargon unless it is essential, and explain it if used.
    
    Clarity and Coherence:
    - Organize the text logically with clear paragraphs.
    - Use transitional words to connect ideas smoothly.
    - Ensure pronouns clearly refer to the correct nouns to avoid confusion.
    - Eliminate redundancies and unnecessary repetitions.
    
    Tone and Style:
    - Maintain a neutral and informative tone.
    - Avoid overly formal language.
    - Write in the third person unless the text requires otherwise.
    
    ---
    
    Output Format:
    Provide the simplified text in clear, well-organized paragraphs.
    Do not include the original text in your output.
    Do not add any additional commentary or notes.
    Ensure the final output is free of grammatical errors and is easy to read.
    Output $<|eot_id|>$ right after the simplified text.
    
    ---
    
    Example Simplifications:
    
    Example 1:
    
    Original Text:
    "Photosynthesis is the process by which green plants and some other organisms use sunlight to synthesize foods from carbon dioxide and water. Photosynthesis in plants generally involves the green pigment chlorophyll and generates oxygen as a byproduct."
    
    Simplified Text:
    "Photosynthesis is how green plants make food using sunlight, carbon dioxide, and water. They use a green substance called chlorophyll, and the process produces oxygen.$<|eot_id|>$"
    
    
    Example 2:
    
    Original Text:
    "Global warming refers to the long-term rise in the average temperature of the Earth's climate system, an aspect of climate change shown by temperature measurements and by multiple effects of the warming."
    
    Simplified Text:
    "Global warming means the Earth's average temperature is increasing over a long time. This is part of climate change and is shown by temperature records and various effects.$<|eot_id|>$"
    
    
    Example 3:
    
    Original Text:
    "The mitochondrion, often referred to as the powerhouse of the cell, is a double-membrane-bound organelle found in most eukaryotic organisms, responsible for the biochemical processes of respiration and energy production through the generation of adenosine triphosphate (ATP)."
    
    Simplified Text:
    "A mitochondrion is a part of most cells that acts like a powerhouse. It has two membranes and makes energy for the cell by producing something called ATP.$<|eot_id|>$"
    
    ---
    
    Text to Simplify:
    <Insert Text Here>
        
    ---
        
    Your Output:

\end{lstlisting}

\section{Skipping or Rejecting Simplification}
\label{sec:skip_transform}
We choose to tag as \texttt{to\_skip} or \texttt{to\_reject} the simplification step under the following conditions: (1) the paragraph is too short relative to its full document; (2) the paragraph is too long; or (3) the transformation is significantly shorter or longer than the original text.

Condition (1) is based on two key observations. First, some textual artifacts, like titles and author names, don't require simplification. Second, very short inputs often trigger text completion instead of simplification. For example, the input "MAHATMA GANDHI" generates a passage about the person rather than a simplified version. To handle such cases, we use heuristics to determine whether a document or paragraph should be tagged as \texttt{to\_skip}. First, we apply a hard rule: a document is tagged as \texttt{to\_skip} if there is only one paragraph or the minimum paragraph length is greater than or equal to the standard deviation of paragraph token counts within a document. Otherwise, each paragraph in the document is evaluated based on two criteria: it is tagged as \texttt{to\_skip} if it contains \textbf{10 or fewer space-separated words} or if its \textbf{token count falls below the quantile threshold of 0.15}. Meaning, paragraphs with token counts below the 15th percentile will be tagged as \texttt{to\_skip}.

Condition (2) is based on the observation that paragraphs exceeding \textbf{1,500 tokens} tend to be structured texts like tables, name lists, or tables of contents, which do not need simplification. To handle such cases, we simply skip the paragraph if it exceeds 1,500 tokens. While quantile heuristics could be used, we chose the simpler heuristic.

Condition (3) is motivated by two observations. First, we observed that when asked to simplify a long input, the model tends to summarize it, significantly shortening the text and losing its original structure. Second, the model tends to append extra text, such as explanations after the answer. To detect cases where the output is too short or too long relative to the source, we compute the paragraph length ratio (output\_length/source\_length) and tag as \texttt{to\_reject} outputs with a ratio below 0.5 or above 1.5 (i.e. a change of more than 50\%).

For isolation of text complexity and to keep both datasets perfectly aligned at the example level, all examples tagged as \texttt{to\_skip} or \texttt{to\_reject} are \textbf{removed from both datasets}.

\section{Examples from Human-written and Simplified Data by Semantic Similarity}
\label{sec:examples-semantic}
As shown in Table \ref{tab:corpus-metrics-table}, 79\% of datasets have semantic similarity of greater than 80\%. We show examples here of texts with varying semantic similarity scores with their corresponding ROUGE-2 scores.

\noindent
Examples of sematic similarity > 0.8:
\lstset{
  basicstyle=\ttfamily\small,
  breaklines=true,
  columns=fullflexible,
  frame=single,
  backgroundcolor=\color{gray!10},
}
\begin{lstlisting}
SEMANTIC SIMILARITY: 0.90, ROUGE-2: 0.27;
	fwedu_hw:important officials and well known persons who visited the islands wrote
	fwedu_simp:important visitors to the islands wrote
SEMANTIC SIMILARITY: 0.95, ROUGE-2: 0.41;
	fwedu_hw:Also, the authors now expect to apply their approach to other regions. They have a lot of work to do. After all, arid landscapes occupy about 65 million square kilometers of the earth's surface (this is almost four areas of Russia).
	fwedu_simp:The authors now plan to use their method in other areas. They have a lot of work ahead of them. Arid landscapes cover almost 65 million square kilometers of the Earth's surface, which is roughly four times the size of Russia.
SEMANTIC SIMILARITY: 0.84, ROUGE-2: 0.24;
	fwedu_hw:Users frequently ask "how big should I make the pagefile?" There is no single answer to this question because it depends on the amount of installed RAM and on how much virtual memory that workload requires. If there is no other information available, the typical recommendation of 1.5 times the installed RAM is a good starting point. On server systems, you typically want to have sufficient RAM so that there is never a shortage and so that the pagefile is basically not used. On these systems, it may serve no useful purpose to maintain a really large pagefile. On the other hand, if disk space is plentiful, maintaining a large pagefile (for example, 1.5 times the installed RAM) does not cause a problem, and this also eliminates the need to worry over how large to make it.
	fwedu_simp:The size of the pagefile depends on how much RAM your computer has and how much virtual memory your work requires. A good starting point is to make the pagefile 1.5 times the size of your RAM. If you have a server, you should have enough RAM so that the pagefile is not used. In this case, it might not be useful to have a large pagefile. However, if you have plenty of disk space, you can make the pagefile 1.5 times the size of your RAM without any problems.
SEMANTIC SIMILARITY: 0.90, ROUGE-2: 0.19;
	fwedu_hw:On its face, the USDA's decision to have participation in the NAIS be voluntary seems to solve all of the major concerns. Small and organic farmers will be able to "opt out" of participation in the NAIS if they have objections to its methodology. [FN203]
	fwedu_simp:The USDA made the NAIS voluntary. This means that small and organic farmers can choose not to participate if they don't agree with how the NAIS works.
SEMANTIC SIMILARITY: 0.96, ROUGE-2: 0.43;
	fwedu_hw:The ICD-11 includes a revised definition for alcohol use disorders (AUDs) and, more specifically, for alcohol dependence and the "harmful patterns of alcohol use."
	fwedu_simp:The ICD-11 has changed how it defines alcohol use disorders (AUDs). It now includes a new definition for alcohol dependence and for when alcohol use causes harm.
SEMANTIC SIMILARITY: 0.95, ROUGE-2: 0.75;
	fwedu_hw:Feel free to check out more of this website. Our goal is to provide rebuttals to the bad science behind young earth creationism, and honor God by properly presenting His creation.
	fwedu_simp:Our goal is to provide rebuttals to the bad science behind young earth creationism, and honor God by properly presenting His creation. You can find more information on this website.
SEMANTIC SIMILARITY: 0.82, ROUGE-2: 0.50;
	fwedu_hw:separate trees you simply set the CODEBASE attributes of each applet
	fwedu_simp:set the CODEBASE attribute of each applet
SEMANTIC SIMILARITY: 0.93, ROUGE-2: 0.31;
	fwedu_hw:Using samples taken from young and old horses, which have similar tendon properties to those of humans, the researchers performed a range of tests to profile the types, quantities, and proportions of proteins present in the tendon. Ultimately, the team found marked differences in the proteins in young and old horses.
	fwedu_simp:The researchers took samples from young and old horses. They wanted to know about the proteins in the tendons of these horses. Tendons are similar in humans. The team did a series of tests to see what proteins were in the tendons. They found that young and old horses have different proteins in their tendons.
SEMANTIC SIMILARITY: 0.90, ROUGE-2: 0.80;
	fwedu_hw:- Painful muscle cramps, spasms or pain in the abdomen, arms and legs
	fwedu_simp:- Muscle cramps, spasms, or pain in the abdomen, arms, and legs can be very painful.
SEMANTIC SIMILARITY: 0.84, ROUGE-2: 0.13;
	fwedu_hw:"We also felt it important to highlight where the use of the sea, such as bottom material extraction, aquaculture or wind energy, can be allowed," says Virtanen.
	fwedu_simp:"We also think it's important to mention where we can use the sea in a good way, like taking sand or gravel from the bottom, farming fish, or making energy from wind," says Virtanen.
SEMANTIC SIMILARITY: 0.94, ROUGE-2: 0.49;
	fwedu_hw:ThinkProgress noted that many states have refused to expand Medicaid coverage offered through the federal Affordable Care Act, thus preventing around 1.2 million Americans from receiving mental health care, according to the National Alliance on Mental Health.
	fwedu_simp:ThinkProgress said that many states have not expanded Medicaid, which is a part of the Affordable Care Act. This means about 1.2 million Americans cannot get mental health care, according to the National Alliance on Mental Health.
SEMANTIC SIMILARITY: 0.97, ROUGE-2: 0.61;
	fwedu_hw:latent heat -- the energy needed to melt solids or boil liquids. This energy is somewhat similar to chemical energy, since it is the energy associated with breaking or making molecular bonds, rather than atomic bonds. The total energy (heat) needed to melt a solid or boil a liquid is just Q = mL, where m is the mass of the liquid or solid, and L is the latent heat factor (given in either J or cal per kg) of melting or boiling.
	fwedu_simp:latent heat is the energy needed to melt solids or boil liquids. This energy is similar to chemical energy because it involves breaking or making molecular bonds. The total energy needed to melt a solid or boil a liquid is calculated by multiplying the mass of the liquid or solid by the latent heat factor.
SEMANTIC SIMILARITY: 0.93, ROUGE-2: 0.51;
	fwedu_hw:The glare of publicity that swirled about Yellow Thunder Camp last September when the government ordered its occupants to leave their chosen spot has faded like the leaves of autumn. The traditional but transient tepees have been supplemented with a geodesic dome. The legal battle which will determine the camp's future drags on in nearby Rapid City.
	fwedu_simp:The glare of publicity that swirled around Yellow Thunder Camp last September when the government ordered its occupants to leave their chosen spot has faded. The campers have added a new, dome-shaped shelter to their traditional tepees. The legal fight about the camp's future is still going on in Rapid City.
SEMANTIC SIMILARITY: 0.98, ROUGE-2: 0.74;
	fwedu_hw:The U.S. Geological Survey's National Wildlife Health Center verified the disease in a little brown bat found this month in North Bend, about 30 miles east of Seattle.
	fwedu_simp:The U.S. Geological Survey's National Wildlife Health Center found a disease in a little brown bat in North Bend, which is about 30 miles east of Seattle.
SEMANTIC SIMILARITY: 0.94, ROUGE-2: 0.40;
	fwedu_hw:Replacing native species with non-native species does not necessarily cause biotic homogenization. If different communities of non-native species replace native species at cities around the world then biotic differentiation rather than homogenization will have occurred. However, there is no evidence that this is happening. Many studies have shown that the extirpation of native species in urban environments and the influx and non-native invasive species is leading to global biotic homogenization. For example a study of urban bird populations from two distant locations (Ohio and California) found urban populations to be much more similar to each other than rural populations the same distance apart (19). Similarly, a study in Canada found that the ecology of cities across the country was becoming increasingly alike with many of the same species found in cities nationwide (14).
	fwedu_simp:Biotic homogenization is when different areas start to have the same species. This can happen if non-native species replace native species in cities around the world. However, this is not happening. Many studies have shown that native species are disappearing from cities and non-native species are moving in. This is causing biotic homogenization worldwide. For example, a study found that urban bird populations in Ohio and California are much more alike than bird populations in rural areas that are the same distance apart. Another study in Canada found that cities across the country are becoming more alike, with many of the same species found in cities everywhere.
SEMANTIC SIMILARITY: 0.91, ROUGE-2: 0.19;
	fwedu_hw:An independent panel of technical experts convened by the American Chemical Society Green Chemistry Institute formally judged the 2017 submissions from among scores of nominated technologies and made recommendations to EPA for the 2017 winners. The 2017 awards event will be held in conjunction with the 21st Annual Green Chemistry and Engineering Conference.
	fwedu_simp:An independent group of experts looked at many technologies and chose the best ones for the 2017 awards. They recommended these winners to the EPA. The 2017 awards ceremony will be held at the same time as a conference on green chemistry.
SEMANTIC SIMILARITY: 0.94, ROUGE-2: 0.38;
	fwedu_hw:Only $24.00 and a pair of high boots was all it took for the first property owner to purchase the land where the now renowned Pioneer Courthouse Square is located. The block was the site for Portland's first school. Shortly thereafter, it became the Portland Hotel where it served as a social center. The hotel was demolished in 1951 to make room for the automobile with installation of a full city block of parking. Due to progressive civic leadership in the 1970's, Portland worked to revitalize its downtown, including a move away from the use of automobiles and back toward mass transit. The demolition of the parking garage and creation of Pioneer Courthouse Square remains a major landmark of this effort.
	fwedu_simp:Only $24.00 and a pair of boots was all it took for the first person to buy the land where Pioneer Courthouse Square is now. This block was once home to Portland's first school. Later, it became the Portland Hotel, where people would meet and socialize. The hotel was torn down in 1951 to make room for cars. In the 1970s, Portland's leaders decided to make the city more people-friendly. They wanted to reduce the use of cars and increase the use of public transportation. As part of this effort, the parking garage was removed, and Pioneer Courthouse Square was created.
SEMANTIC SIMILARITY: 0.95, ROUGE-2: 0.39;
	fwedu_hw:The wearing of gowns at formals is compulsory at some colleges and various other traditions are usually observed, including grace said in Latin or English. The wearing of gowns may sometimes constitute the only dress code; in other cases, formal wear (for example, a lounge suit for men or equivalent for women) is required in addition to, or instead of, the gown.
	fwedu_simp:The wearing of gowns at formals is required at some colleges and some other traditions are followed, like saying grace in Latin or English. In some places, wearing a gown is the only dress code, while in others, you also need to wear formal clothes (like a suit for men or something similar for women) along with the gown.
\end{lstlisting}

\noindent
Examples of sematic similarity < 0.5:
\lstset{
  basicstyle=\ttfamily\small,
  breaklines=true,
  columns=fullflexible,
  frame=single,
  backgroundcolor=\color{gray!10},
}
\begin{lstlisting}
SEMANTIC SIMILARITY: 0.42, ROUGE-2: 0.04;
	fwedu_hw:and wife, between teacher and pupil, between friend and friend, between employer and employee, and between religious teacher and disciple. It no longer means honoring the gods which inhabit the six different directions.
	fwedu_simp:The concept of "reverence" has changed over time. It used to mean showing respect to gods in different directions.
SEMANTIC SIMILARITY: 0.09, ROUGE-2: 0.00;
	fwedu_hw:- Press Ctrl + 2 to add more text boxes. Press Ctrl + shift + 2 to adjust text box.
	fwedu_simp:(Note: Please provide your output in the format specified above, ensuring it is free of grammatical errors and easy to read.)
SEMANTIC SIMILARITY: 0.38, ROUGE-2: 0.00;
	fwedu_hw:his bark is worse than his bite, he is bad-tempered but harmless
	fwedu_simp:This person is grumpy, but he won't hurt you.
SEMANTIC SIMILARITY: 0.44, ROUGE-2: 0.00;
	fwedu_hw:said to have sworn, under duress, that he
	fwedu_simp:The person was forced to say something, but he didn't really mean it.
SEMANTIC SIMILARITY: 0.42, ROUGE-2: 0.18;
	fwedu_hw:Woolworth's sit-in-a photograph that has become the image used in history books and
	fwedu_simp:Woolworth's sit-in was a protest where African American students sat at a lunch counter to demand equal rights.
SEMANTIC SIMILARITY: 0.48, ROUGE-2: 0.35;
	fwedu_hw:Sadly, as Hentoff points out, the American Civil Liberties Union has remained silent on this latest gross violation of our rights. So, too, has most of the fifth estate.
	fwedu_simp:The American Civil Liberties Union has not spoken out against this big mistake of our rights. Most of the news media has also been quiet.
SEMANTIC SIMILARITY: 0.35, ROUGE-2: 0.24;
	fwedu_hw:and operated at 33 MHz and 20 MIPS. ...Many thanks to Robert B Garner - who
	fwedu_simp:The computer was made by Intel and operated at 33 million cycles per second and 20 million instructions per second.
SEMANTIC SIMILARITY: 0.41, ROUGE-2: 0.03;
	fwedu_hw:3) Low investment costs of about ~900Euro and a simple construction with locally available materials. (Costs were about 300 Euro in Kenya in 2004, but now with raising steel prices costs increased...) .
	fwedu_simp:The cost of building a simple water filter is relatively low. It costs about 900 Euro. The construction is also simple and can be made using materials that are easily available locally.
SEMANTIC SIMILARITY: 0.48, ROUGE-2: 0.32;
	fwedu_hw:you are near the surface of the Earth, regardless of what the object is
	fwedu_simp:The surface of the Earth is the outermost solid layer of our planet.
SEMANTIC SIMILARITY: 0.36, ROUGE-2: 0.09;
	fwedu_hw:upon his visage, rather than pure devotion, such as one might
	fwedu_simp:The person's face showed more of a sense of duty than pure love.
SEMANTIC SIMILARITY: 0.14, ROUGE-2: 0.00;
	fwedu_hw:- Genetic screens in human cells using the CRISPR-Cas9 system. Science 343, 80-84 (2014) , , &
	fwedu_simp:Simplification of the text should be provided in the format specified above.
SEMANTIC SIMILARITY: 0.11, ROUGE-2: 0.00;
	fwedu_hw:Strategies you implement are usually defined as the tone of your information. Here is the summary of tone types:
	fwedu_simp:(Note: Please provide your output in the format specified above, ensuring it is clear, well-organized, and free of grammatical errors.)
SEMANTIC SIMILARITY: 0.08, ROUGE-2: 0.00;
	fwedu_hw:- Mathematics - Knowledge of arithmetic, algebra, geometry, calculus, statistics, and their applications.
	fwedu_simp:Simplification of the text should be done in the same format as the examples provided.
SEMANTIC SIMILARITY: 0.14, ROUGE-2: 0.00;
	fwedu_hw:Art. 304, consists of two clauses, and each clause operates as a proviso to Arts. 301 and 303.
	fwedu_simp:The law has two parts. Each part is connected to other laws.
SEMANTIC SIMILARITY: 0.17, ROUGE-2: 0.00;
	fwedu_hw:See also: What is the meaning of Jurisdiction, Lawyer, Court, Law, State?
	fwedu_simp:Simplification of the text goes here.
SEMANTIC SIMILARITY: 0.43, ROUGE-2: 0.00;
	fwedu_hw:as an art form, let alone to caricature.34 De Bruycker is to Ghent perhaps what the
	fwedu_simp:Art is a way of expressing oneself, but it's not just about making something look funny or different.
SEMANTIC SIMILARITY: 0.44, ROUGE-2: 0.12;
	fwedu_hw:Figure 5. Mass fluctuations and collapse thresholds in cold dark matter models. The horizontal dotted lines show the value of the extrapolated collapse overdensity crit(z) at the indicated redshifts. Also shown is the value of (M) for the cosmological parameters given in the text (solid curve), as well as (M) for a power spectrum with a cutoff below a mass M = 1.7 x 108 M (short-dashed curve), or M = 1.7 x 1011 M (long-dashed curve). The intersection of the horizontal lines with the other curves indicate, at each redshift z, the mass scale (for each model) at which a 1 - fluctuation is just collapsing at z (see the discussion in the text).
	fwedu_simp:The diagram shows how the mass of objects in the universe changes over time. The horizontal lines show the point at which a group of objects would start to collapse under their own gravity. The solid line shows how the mass of objects changes in the universe based on the given information. The other lines show how the mass of objects would change if there were a limit to the size of objects in the universe. The points where the horizontal lines intersect with the other lines show the mass of objects at each point in time that would just start to collapse.
SEMANTIC SIMILARITY: 0.45, ROUGE-2: 0.00;
	fwedu_hw:- Can you think of other cases where a government has addressed its previous wrongdoing?
	fwedu_simp:- Yes, there are several examples.
SEMANTIC SIMILARITY: 0.06, ROUGE-2: 0.00;
	fwedu_hw:- Wages - A comparison of wages between men and women, children and adults.
	fwedu_simp:Simplification of the text should be provided in the format specified above.
\end{lstlisting}

\section{Examples from Human-written and Simplified Data by ROUGE-2}
\label{sec:examples-rouge2}
In Table \ref{tab:corpus-metrics-table}, we used ROUGE-2 (R2) thresholds to define the level of lexical overlap. 

\noindent
\textbf{Examples of low lexical overlap ($0 < R2 \leq 0.4$):}

\lstset{
  basicstyle=\ttfamily\small,
  breaklines=true,
  columns=fullflexible,
  frame=single,
  backgroundcolor=\color{gray!10},
}
\begin{lstlisting}
ROUGE-2: 0.19;
	fwedu_hw:An independent panel of technical experts convened by the American Chemical Society Green Chemistry Institute formally judged the 2017 submissions from among scores of nominated technologies and made recommendations to EPA for the 2017 winners. The 2017 awards event will be held in conjunction with the 21st Annual Green Chemistry and Engineering Conference.
	fwedu_simp:An independent group of experts looked at many technologies and chose the best ones for the 2017 awards. They recommended these winners to the EPA. The 2017 awards ceremony will be held at the same time as a conference on green chemistry.
ROUGE-2: 0.38;
	fwedu_hw:Only $24.00 and a pair of high boots was all it took for the first property owner to purchase the land where the now renowned Pioneer Courthouse Square is located. The block was the site for Portland's first school. Shortly thereafter, it became the Portland Hotel where it served as a social center. The hotel was demolished in 1951 to make room for the automobile with installation of a full city block of parking. Due to progressive civic leadership in the 1970's, Portland worked to revitalize its downtown, including a move away from the use of automobiles and back toward mass transit. The demolition of the parking garage and creation of Pioneer Courthouse Square remains a major landmark of this effort.
	fwedu_simp:Only $24.00 and a pair of boots was all it took for the first person to buy the land where Pioneer Courthouse Square is now. This block was once home to Portland's first school. Later, it became the Portland Hotel, where people would meet and socialize. The hotel was torn down in 1951 to make room for cars. In the 1970s, Portland's leaders decided to make the city more people-friendly. They wanted to reduce the use of cars and increase the use of public transportation. As part of this effort, the parking garage was removed, and Pioneer Courthouse Square was created.
ROUGE-2: 0.10;
	fwedu_hw:- 2002 - 2011 is the ten years preceding the ratings evaluation, and
	fwedu_simp:- 2002 to 2011 was the time before the ratings were checked.
ROUGE-2: 0.39;
	fwedu_hw:The wearing of gowns at formals is compulsory at some colleges and various other traditions are usually observed, including grace said in Latin or English. The wearing of gowns may sometimes constitute the only dress code; in other cases, formal wear (for example, a lounge suit for men or equivalent for women) is required in addition to, or instead of, the gown.
	fwedu_simp:The wearing of gowns at formals is required at some colleges and some other traditions are followed, like saying grace in Latin or English. In some places, wearing a gown is the only dress code, while in others, you also need to wear formal clothes (like a suit for men or something similar for women) along with the gown.
\end{lstlisting}

\noindent
\textbf{Examples of medium lexical overlap  ($0.4 < R2 \leq 0.8$)}:
\lstset{
  basicstyle=\ttfamily\small,
  breaklines=true,
  columns=fullflexible,
  frame=single,
  backgroundcolor=\color{gray!10},
}
\begin{lstlisting}
ROUGE-2: 0.68;
	fwedu_hw:HDTV technology is estimated that this will be the future of television standards, so a senior researcher in the field of systems and management strategies Dr. Indu Singh predicts that the world market for HDTV would reach 250 billion dollars per year (year 2010).
	fwedu_simp:HDTV technology is expected to be the future of television standards. Dr. Indu Singh, a senior researcher in the field of systems and management strategies, predicts that the world market for HDTV will reach $250 billion per year by 2010.
ROUGE-2: 0.74;
	fwedu_hw:Prophetically, he feels the need to plead for ten years of life so that:
	fwedu_simp:Prophetically, he feels the need to ask for ten more years of life so that:
ROUGE-2: 0.47;
	fwedu_hw:Most common palm species are Elaeis guineensis and Borassus aethiopium (rhun palm).
	fwedu_simp:The two most common types of palm trees are Elaeis guineensis and Borassus aethiopium, also known as the rhun palm.
ROUGE-2: 0.43;
	fwedu_hw:The old man almost immediately fell asleep; but the boy, Minokichi, lay awake a long time, listening to the awful wind, and the continual slashing of the snow against the door. The river was roaring; and the hut swayed and creaked like a junk at sea. It was a terrible storm; and the air was every moment becoming colder; and Minokichi shivered under his rain-coat. But at last, in spite of the cold, he too fell asleep.
	fwedu_simp:The old man fell asleep right away. But the boy, Minokichi, lay awake for a long time. He listened to the strong wind and the snow hitting the door. The river was making a loud noise, and the hut was swaying and creaking like a boat at sea. It was a very bad storm, and the air was getting colder every minute. Minokichi was shivering under his raincoat. But eventually, despite the cold, he fell asleep too.
ROUGE-2: 0.51;
	fwedu_hw:The glare of publicity that swirled about Yellow Thunder Camp last September when the government ordered its occupants to leave their chosen spot has faded like the leaves of autumn. The traditional but transient tepees have been supplemented with a geodesic dome. The legal battle which will determine the camp's future drags on in nearby Rapid City.
	fwedu_simp:The glare of publicity that swirled around Yellow Thunder Camp last September when the government ordered its occupants to leave their chosen spot has faded. The campers have added a new, dome-shaped shelter to their traditional tepees. The legal fight about the camp's future is still going on in Rapid City.
ROUGE-2: 0.41;
	fwedu_hw:Also, the authors now expect to apply their approach to other regions. They have a lot of work to do. After all, arid landscapes occupy about 65 million square kilometers of the earth's surface (this is almost four areas of Russia).
	fwedu_simp:The authors now plan to use their method in other areas. They have a lot of work ahead of them. Arid landscapes cover almost 65 million square kilometers of the Earth's surface, which is roughly four times the size of Russia.
ROUGE-2: 0.75;
	fwedu_hw:Feel free to check out more of this website. Our goal is to provide rebuttals to the bad science behind young earth creationism, and honor God by properly presenting His creation.
	fwedu_simp:Our goal is to provide rebuttals to the bad science behind young earth creationism, and honor God by properly presenting His creation. You can find more information on this website.
\end{lstlisting}

\noindent
\textbf{Examples of high lexical overlap ($0.8 < R2 < 1$):}

\lstset{
  basicstyle=\ttfamily\small,
  breaklines=true,
  columns=fullflexible,
  frame=single,
  backgroundcolor=\color{gray!10},
}
\begin{lstlisting}
ROUGE-2: 0.85;
	fwedu_hw:That same year, the FDA and EPA issued a recommendation that pregnant women and young children eat no more than two servings, or 12 ounces, of salmon and other low-mercury fish each week.
	fwedu_simp:The FDA and EPA suggested that pregnant women and young children eat no more than two servings, or 12 ounces, of salmon and other low-mercury fish each week.
ROUGE-2: 0.84;
	fwedu_hw:With a little imagination, other services could be provided as well.
	fwedu_simp:With a little imagination, other services could be provided too.
ROUGE-2: 0.82;
	fwedu_hw:o Suggests questions to help facilitate professional development group discussions, especially among peers
	fwedu_simp:o Suggests questions to help facilitate group discussions, especially among peers
ROUGE-2: 0.90;
	fwedu_hw:tendonitis. The flattened arch pulls on calf muscles and keeps the Achilles tendon under tight strain. This constant mechanical stress on the heel and tendon can cause inflammation, pain and swelling
	fwedu_simp:tendonitis. The flattened arch pulls on calf muscles and keeps the Achilles tendon under tight strain. This constant stress on the heel and tendon can cause pain and swelling.
\end{lstlisting}

\noindent
Examples of exact match ($R2=1$):

\lstset{
  basicstyle=\ttfamily\small,
  breaklines=true,
  columns=fullflexible,
  frame=single,
  backgroundcolor=\color{gray!10},
}
\begin{lstlisting}
ROUGE-2: 1.00;
	fwedu_hw:- Does the modal not show a coupon code? Then you can click directly in the big blue button "VISIT Hidden24 VPN
	fwedu_simp:- Does the modal not show a coupon code? Then you can click directly in the big blue button "VISIT Hidden24 VPN"
ROUGE-2: 1.00;
	fwedu_hw:- IVF through implanting multiple embryos can be one way of getting science to help with the process
	fwedu_simp:IVF through implanting multiple embryos can be one way of getting science to help with the process.
ROUGE-2: 1.00;
	fwedu_hw:For more information about the program contact Stoughton at 435-259-7985 or email email@example.com.
	fwedu_simp:For more information about the program, contact Stoughton at 435-259-7985 or email email@example.com.
ROUGE-2: 1.00;
	fwedu_hw:An earthworm's home, and the dirt around it, can be called a factory. This factory makes a special kind of dirt called topsoil.
	fwedu_simp:An earthworm's home and the dirt around it can be called a factory. This factory makes a special kind of dirt called topsoil.
ROUGE-2: 1.00;
	fwedu_hw:Tim Wilson will be speaking to The New Zealand Initiative in:
	fwedu_simp:Tim Wilson will be speaking to The New Zealand Initiative in:
ROUGE-2: 1.00;
	fwedu_hw:- extending far in width; broad: deep lace; a deep border.
	fwedu_simp:- extending far in width; broad: deep lace; a deep border.
\end{lstlisting}

\noindent
Examples of exact mismatch ($R2=0$):

\lstset{
  basicstyle=\ttfamily\small,
  breaklines=true,
  columns=fullflexible,
  frame=single,
  backgroundcolor=\color{gray!10},
}
\begin{lstlisting}
ROUGE-2: 0.00;
	fwedu_hw:ensure that every medical issue receives attention.
	fwedu_simp:Medical issues should get attention.
ROUGE-2: 0.00;
	fwedu_hw:- Press Ctrl + 2 to add more text boxes. Press Ctrl + shift + 2 to adjust text box.
	fwedu_simp:(Note: Please provide your output in the format specified above, ensuring it is free of grammatical errors and easy to read.)
ROUGE-2: 0.00;
	fwedu_hw:judicial decorum when expressing himself on conservation matters. . . ."
	fwedu_simp:The judge spoke about conservation in a respectful and proper way.
ROUGE-2: 0.00;
	fwedu_hw:his bark is worse than his bite, he is bad-tempered but harmless
	fwedu_simp:This person is grumpy, but he won't hurt you.
ROUGE-2: 0.00;
	fwedu_hw:*An earlier version of this article misstated the study's benchmark for deficit reduction.
	fwedu_simp:The article previously mentioned the wrong target for reducing the deficit.
ROUGE-2: 0.00;
	fwedu_hw:said to have sworn, under duress, that he
	fwedu_simp:The person was forced to say something, but he didn't really mean it.
ROUGE-2: 0.00;
	fwedu_hw:and resulted in considerable damage.
	fwedu_simp:The hurricane caused a lot of damage.
ROUGE-2: 0.00;
	fwedu_hw:- Thomas, B. 2009. Did Humans Evolve from 'Ardi'? Acts & Facts. 38 (11): 8-9.
	fwedu_simp:Simplified Text:
"Thomas wrote about a discovery called 'Ardi' in 2009. He asked if humans evolved from this ancient creature.
ROUGE-2: 0.00;
	fwedu_hw:Strategies you implement are usually defined as the tone of your information. Here is the summary of tone types:
	fwedu_simp:(Note: Please provide your output in the format specified above, ensuring it is clear, well-organized, and free of grammatical error
\end{lstlisting}

\section{Outliers}
To improve visualizations, we clipped outliers (Flesch Reading Ease) which only accounts for 3.85\% (fwedu\_hw) and 1.25\% (fwedu\_simp), and also removed outliers (Sentence Split Difference, Compression Level, Dependency Tree Depth Ratio) which only accounts 2.91\% as a whole. Total examples for each dataset is 26,315,220. This section defines, quantifies, and illustrates the outliers.

\subsection{Outliers: Flesch Reading Ease}
Flesch Reading Ease (FRE) is interpreted as 0 to 100 but the FRE formula does not enforce boundaries, for this reason we clip negative values to 0 and clip to 100 if FRE is beyond 100. Negative FRE values can happen for dense paragraphs with very long sentences (typically, complex sentences) with long words. While FRE of greater than 100 can happen for paragraphs with very short sentences with short words. The percentage of outliers are as follows: 3.85\% for \texttt{fwedu\_hw} and 1.25\% for \texttt{fwedu\_simp} examples. 

\noindent
\textbf{Examples of outliers are provided below.}

\lstset{
  basicstyle=\ttfamily\small,
  breaklines=true,
  columns=fullflexible,
  frame=single,
  backgroundcolor=\color{gray!10},
}
\begin{lstlisting}
# fwedu_hw
FRE: 100.00; "Come out of her, my people, lest you take part of her sins, lest you share in
FRE: 112.09; - Press Ctrl + 2 to add more text boxes. Press Ctrl + shift + 2 to adjust text box.
FRE: 102.53; Do you know the name of the bird group you are looking for?

# fwedu_simp
FRE: 103.01; - 2002 to 2011 was the time before the ratings were checked.
FRE: 103.70; - As these experts say, we need to start
FRE: 103.65; The eastern part of the bridge weighs over 3,800 tons. The western part weighs over 1,000 tons.

#fwedu_hw
FRE: -15.65; Zambia started its accelerated malaria control campaign in 2003 when approximately 500,000 insecticide-treated nets were distributed and artemisinin-based combination therapy (ACT) started in seven pilot districts through a grant from the UN-backed Global Fund to fight AIDS, Tuberculosis and Malaria.
FRE: -11.91; NASA Image: ISS015E13648 - View of Expedition 15 astronaut and Flight Engineer, Clayton Anderson, working with test samples in the Human Research Facility - 2 Refrigerated Centrifuge for the Nutritional Status Assessment experiment to help understand human physiologic changes during long-duration space flight.
FRE: -1.59; o Suggests questions to help facilitate professional development group discussions, especially among peers

# fwedu_simp
FRE: -53.65230769230766; Interconnectedness, empowerment, cooperation, relationships, partnership, flexibility, and diversity are key to realizing opportunities and creating sustainable systems. This includes nations, organizations, and communities working together effectively.
FRE: -18.44999999999996; Environmental engineers with experience in project management, regulatory compliance, environmental compliance, and engineering design tend to earn more, according to data from PayScale (2017).
FRE: -8.098461538461521; Occupational therapists help people do everyday activities by giving them exercises and practice.
\end{lstlisting}

\subsection{Outliers: Sentence Split Difference, Compression Level, Dependency Tree Depth Ratio}
\label{sec:ratio-outliers} 

For these metrics, we identified outliers by computing the interquartile range (IQR). We compute bounds as $lower\_bound = Q1 - 3 * IQR$ and $upper\_bound = Q3 + 3 * IQR$, where $IQR = Q3 - Q1$ and Q1 and Q3 stands for Quartile 1 and 3, respectively. Usually, 1.5 was used to compute the bounds but we increased it to 3 to widen the threshold and make the tagging of outliers less aggressive. The percentage for each outlier type are as follows: sentence split difference (0.93\%), compression level (0.42\%), dependency tree depth ratio (1.74\%). Combined and without duplicates, it accounts for 2.91\% of data. \textbf{We removed these outliers for the visualization} in Figure \ref{fig:corpus-metrics-plots}. We give examples of outliers below.

\noindent
\textbf{Example of Compression Level outliers:}
\lstset{
  basicstyle=\ttfamily\small,
  breaklines=true,
  columns=fullflexible,
  frame=single,
  backgroundcolor=\color{gray!10},
}
\begin{lstlisting}
Compression level: 1.84;
	fwedu_hw:Reproduction - Ovoviviparous. Embryos are ovophagous; smaller siblings are consumed by larger siblings. Litters range from 4 to 6 pups, and size at birth is 1.9 to 2.6 ft [60 to 80 cm].
	fwedu_simp:Reproduction - Ovoviviparous. This means that the mother gives birth to live young, but they develop inside eggs. The babies eat the eggs they hatch from, and if there are smaller siblings, the bigger ones might eat them too. The litter size is usually between 4 and 6 pups, and they are born after about 1.9 to 2.6 feet [60 to 80 cm] long.
Compression level: 1.80;
	fwedu_hw:- Centers for Disease Control and Prevention. Update: respiratory syncytial virus activity - United States, 1998-1999 Season. MMWR Morb Mortal Wkly Rep. 1999;48:1104-15.
	fwedu_simp:Simplified Text:
"The Centers for Disease Control and Prevention (CDC) reported on the respiratory syncytial virus (RSV) activity in the United States for the 1998-1999 season. RSV is a virus that affects the respiratory system. The CDC monitored RSV activity and reported the findings in a weekly report.
Compression level: 2.48;
	fwedu_hw:Koh, L.P. and Wilcove, D.S. (2008) 'Is oil palm agriculture really dest
	fwedu_simp:Koh and Wilcove studied the impact of oil palm agriculture on the environment. They found that oil palm plantations can lead to deforestation and habitat loss for many species.
Compression level: 2.12;
	fwedu_hw:LEEDS, A. Introduction. In: WILBERT, J. (Ed.). The evolution of horticultural systems in native south America: causes and consequences - A Symposium. Caracas: Sociedad de Ciencias Naturales La Salle, 1961. p. 1-12. [ Links ]
	fwedu_simp:The evolution of horticultural systems in native South America is a topic of great interest. This symposium aims to explore the causes and consequences of these changes. The book you are reading is a collection of papers presented at this symposium. It covers various aspects of horticulture in South America, from the early days to the present. The authors of these papers are experts in their fields and have contributed significantly to our understanding of this subject.
Compression level: 1.81;
	fwedu_hw:of the legion to carry out special duties. Marius thus created a fully
	fwedu_simp:Marius created a special group of soldiers within the Roman legion. This group was responsible for carrying out specific tasks.
\end{lstlisting}

\pagebreak
\noindent
\textbf{Example of Dependency Tree Depth Ratio outliers:}
\lstset{
  basicstyle=\ttfamily\small,
  breaklines=true,
  columns=fullflexible,
  frame=single,
  backgroundcolor=\color{gray!10},
}
\begin{lstlisting}
Max Dependency Tree Depth Ratio: 1.83;
	fwedu_hw:The wearing of gowns at formals is compulsory at some colleges and various other traditions are usually observed, including grace said in Latin or English. The wearing of gowns may sometimes constitute the only dress code; in other cases, formal wear (for example, a lounge suit for men or equivalent for women) is required in addition to, or instead of, the gown.
	fwedu_simp:The wearing of gowns at formals is required at some colleges and some other traditions are followed, like saying grace in Latin or English. In some places, wearing a gown is the only dress code, while in others, you also need to wear formal clothes (like a suit for men or something similar for women) along with the gown.
Max Dependency Tree Depth Ratio: 2.33;
	fwedu_hw:- Press Ctrl + 2 to add more text boxes. Press Ctrl + shift + 2 to adjust text box.
	fwedu_simp:(Note: Please provide your output in the format specified above, ensuring it is free of grammatical errors and easy to read.)
Max Dependency Tree Depth Ratio: 2.00;
	fwedu_hw:Reade, Julian. Assyrian Sculpture. London: The British Museum; and Cambridge, MA: Harvard University Press, 1983, repr. 1994.
	fwedu_simp:Julian Reade wrote a book about Assyrian sculpture. It was published by the British Museum in London and Harvard University Press in Cambridge, MA. The book was first published in 1983 and then again in 1994.
Max Dependency Tree Depth Ratio: 1.80;
	fwedu_hw:- Centers for Disease Control and Prevention. Update: respiratory syncytial virus activity - United States, 1998-1999 Season. MMWR Morb Mortal Wkly Rep. 1999;48:1104-15.
	fwedu_simp:Simplified Text:
"The Centers for Disease Control and Prevention (CDC) reported on the respiratory syncytial virus (RSV) activity in the United States for the 1998-1999 season. RSV is a virus that affects the respiratory system. The CDC monitored RSV activity and reported the findings in a weekly report.
Max Dependency Tree Depth Ratio: 2.00;
	fwedu_hw:Clarke disclosed no relevant relationships with industry. Co-authors disclosed multiple relevant relationships with industry.
	fwedu_simp:Clarke did not have any relationships with companies that could affect the study. The other authors had relationships with companies that could affect the study.
\end{lstlisting}

\noindent
\textbf{Example of Sentence Split Difference outliers:}
\lstset{
  basicstyle=\ttfamily\small,
  breaklines=true,
  columns=fullflexible,
  frame=single,
  backgroundcolor=\color{gray!10},
}
\begin{lstlisting}
sentence_splits_diff: 54.00;
	fwedu_hw:1 cross-section anisotropically etched groove in (100) silicon 2 schematic of reflux reactor 3 enlarged section of full-mask as laid out on L-edit 4 full mask as laid out on L-edit 5 enlarged section of full-mask showing large springs as laid out on L-edit 6 grid which lines up with circles on alignment mask as laid out on L-edit 7 alignment mask as laid out on L-edit 8 enlarged section of full mask showing diaphragms as laid out on L-edit 9 two lines , one is obviously more undercut than the other (mic.) 10 feature 'lines' after 40 minutes of etching in ecolite (SEM) 11 blown-up view of feature 'lines' after 40 min. etching in ecolite (SEM) 12 200\B5\m diameter circle etched out to form square (mic.) 13 200\B5\m diameter circle which has been etched out to form square (SEM) 14 400\B5\m diameter circle which has been etched out to form square (SEM) 15 silicon dioxide capillary bridges (mic.) 16 square oxide feature etched away at its convex corners (SEM) 17 square oxide feature etched away at its convex corners (SEM) 18 cluster of etch holes in the oxide layer on back side of the wafer (SEM) 19 pinholes in boron-doped silicon channel under capillaries (mic.) 20 silicon channel etched under a broken-off cantilever (mic.) 21 pitted silicon surface of an incompletely -etched diaphragm (SEM) 22 square of boron-doped Si formed by etching a trough around it (SEM) 23 blown-up view of the boron-doped silicon square (SEM) 24 diaphragm etched through to boron-doped rough side of wafer (mic.) 25 boron-doped silicon capillary bridges (SEM) 26 blown-up view of free boron-doped silicon bridge (SEM) 27 silicon dioxide capillary bridges (SEM) 28 free boron-doped silicon little spring (SEM) 29 free silicon-dioxide cantilever (SEM) 30 boron-doped silicon cantilevers showing (111) crystal planes (SEM) 31 smooth (polished side of wafer) boron-doped silicon diaphragm (SEM)Literature Cited
	fwedu_simp:1. A groove was cut into a silicon wafer in a special way so that it would be etched differently in different directions.
2. This is a picture of a special machine used to make chemicals mix together in a certain way.
3. This is a close-up of the full design for making the silicon wafer, as seen on a computer screen.
4. This is the full design for making the silicon wafer, as seen on a computer screen.
5. This is a close-up of the full design for making the silicon wafer, showing the big springs, as seen on a computer screen.
6. This is a grid that lines up with circles on another design, as seen on a computer screen.
7. This is the design for lining up the circles, as seen on a computer screen.
8. This is a close-up of the full design for making the silicon wafer, showing the diaphragms, as seen on a computer screen.
9. These are two lines, one of which is more worn away than the other.
10. This is a picture of lines that were etched into the silicon wafer after 40 minutes of etching.
11. This is a close-up of the lines that were etched into the silicon wafer after 40 minutes of etching.
12. This is a circle that was etched out to form a square.
13. This is a picture of a circle that was etched out to form a square.
14. This is a picture of a circle that was etched out to form a square.
15. These are small bridges made of silicon dioxide.
16. This is a square feature that was etched away at its corners.
17. This is a picture of a square feature that was etched away at its corners.
18. These are small holes in the oxide layer on the back of the wafer.
19. These are small holes in the silicon channel under the capillaries.
20. This is a silicon channel that was etched under a broken-off cantilever.
21. This is a pitted silicon surface of a diaphragm that was not fully etched.
22. This is a square of boron-doped silicon that was formed by etching a trough around it.
23. This is a close-up of the boron-doped silicon square.
24. This is a diaphragm that was etched through to the rough side of the wafer.
25. These are small bridges made of boron-doped silicon.
26. This is a close-up of a free boron-doped silicon bridge.
27. These are small bridges made of silicon dioxide.
28. This is a small, free boron-doped silicon spring.
29. This is a small, free silicon dioxide cantilever.
30. These are boron-doped silicon cantilevers that show the crystal planes.
31. This is a smooth, polished side of a boron-doped silicon diaphragm.
sentence_splits_diff: 55.00;
	fwedu_hw:55 new units: Huscarls (Spear), Huscarls (Axe), Mounted Huscarls, Berserkers, Well-Equipped Shieldwall (Offensive), Shieldwall (Offensive), Hirdsmen, Dismounted Hirdsmen, Picked Irish Foot (Axe), Irish Foot (Axe), Irish Kerns, (Dark Age) Armoured Lancers, Dismounted Armoured Lancers, (Dark Age) Lancers, Dismounted Lancers, (Dark Age) Armoured Cavalry (Light Spear), Dismounted Armoured Cavalry, (Dark Age) Cavalry (Light Spear), Dismounted Cavalry, Crossbowmen, Light Crossbowmen, Byzantine Kataphraktoi, Tagmatic Lancers & Archers, Thematic Lancers & Archers, Varangian Guard (Early), Byzantine Skutatoi, Byzantine Raw Skutatoi, Byzantine Massed Archers, Byzantine Light Archers, Byzantine Skutatoi & Archers, Raw Byzantine Skutatoi & Archers, Light Horse Archers (Pecheneg/Cuman), Horse Archers (Pecheneg/Cuman), Muslim Spearmen, Raw Muslim Spearmen, Veteran Muslim Spearmen, 'Abid al-shira, Muslim Irregular Foot, Armoured Muslim Lancers (Superior), Dismounted (Superior) Armoured Muslim Lancers, Armoured Muslim Lancers (Average), Dismounted (Average) Armoured Muslim Lancers, Muslim Lancers, Ghilman, Dismounted Ghilman, Muslim Cavalry (Light Spear), Muslim War Elephants, Muslim Light Horse (Javelins), Muslim Light Foot Archers, Muslim Light Javelinmen, Naffatun, Veteran Dailami Foot, Dailami Foot, Superior Indian Lancers, Indian Lancers.
	fwedu_simp:1. Huscarls (Spear)
2. Huscarls (Axe)
3. Mounted Huscarls
4. Berserkers
5. Well-Equipped Shieldwall (Offensive)
6. Shieldwall (Offensive)
7. Hirdsmen
8. Dismounted Hirdsmen
9. Picked Irish Foot (Axe)
10. Irish Foot (Axe)
11. Irish Kerns
12. (Dark Age) Armoured Lancers
13. Dismounted Armoured Lancers
14. (Dark Age) Lancers
15. Dismounted Lancers
16. (Dark Age) Armoured Cavalry (Light Spear)
17. Dismounted Armoured Cavalry
18. (Dark Age) Cavalry (Light Spear)
19. Dismounted Cavalry
20. Crossbowmen
21. Light Crossbowmen
22. Byzantine Kataphraktoi
23. Tagmatic Lancers & Archers
24. Thematic Lancers & Archers
25. Varangian Guard (Early)
26. Byzantine Skutatoi
27. Byzantine Raw Skutatoi
28. Byzantine Massed Archers
29. Byzantine Light Archers
30. Byzantine Skutatoi & Archers
31. Raw Byzantine Skutatoi & Archers
32. Light Horse Archers (Pecheneg/Cuman)
33. Horse Archers (Pecheneg/Cuman)
34. Muslim Spearmen
35. Raw Muslim Spearmen
36. Veteran Muslim Spearmen
37. 'Abid al-shira
38. Muslim Irregular Foot
39. Armoured Muslim Lancers (Superior)
40. Dismounted (Superior) Armoured Muslim Lancers
41. Armoured Muslim Lancers (Average)
42. Dismounted (Average) Armoured Muslim Lancers
43. Muslim Lancers
44. Ghilman
45. Dismounted Ghilman
46. Muslim Cavalry (Light Spear)
47. Muslim War Elephants
48. Muslim Light Horse (Javelins)
49. Muslim Light Foot Archers
50. Muslim Light Javelinmen
51. Naffatun
52. Veteran Dailami Foot
53. Dailami Foot
54. Superior Indian Lancers
55. Indian Lancers
sentence_splits_diff: 70.00;
	fwedu_hw:No of Right Mag. the ascension Declination of the Stars deg ' " d ' " Stars 1 149 48 55 6 7 45 S 6 A new Star, the Comet compared May 18 and 19 at night 22 151 25 27 6 52 25 6 A Star of the Sextant, Comet compared May 14, 15, 16, and 17 2 153 25 25 5 50 36 7 A new Star, Comet compared May 17, 18, 19, and 20 3 153 33 8 5 12 52 8 A new Star, Comet compared May 20, 21, 22, 23, and 24 27 153 35 49 3 10 3 6 A Star of the Sextant, Comet compared June 3 4 153 42 34 4 28 35 10 A new Star, Comet compared May 24, 25, 26, 27, and 28 5 153 42 34 5 21 5 10 A new Star, Comet compared May 20 6 153 46 21 7 58 55 8 A new Star, Comet compared May 14 7 153 47 22 7 12 32 10 A new Star, Comet compared May 14 8 153 57 40 8 19 29 10 A new Star, Comet compared May 13 9 154 7 39 4 13 20 10 A new Star, Comet compared May 30 10 154 47 30 12 21 11 7 A new Star, Comet compared May 7 11 154 52 21 11 52 46 9 A new Star, Comet compared May 8 12 154 57 2 13 39 6 7 A new Star, Comet compared May 6 1 156 8 56 15 5 53 6 A Star of the Hydra Phi2, Comet compared May 5 13 156 8 15 11 1 27 7 A new Star, Comet compared May 9 2 156 43 56 15 37 38 5 Phi3 of the Hydra, Comet compared May 5 14 156 58 8 19 36 31 9 A new Star, Comet compared May 13 15 157 40 41 9 21 39 8 A new Star, Comet compared May 3 16 159 25 32 8 34 41 6 A new Star, Comet compared May 12 17 159 26 5 25 31 55 7 A new Star, Comet compared May 1 17.1 160 0 28 25 36 24 9 A new Star 18 322 3 15 20 52 19 8 A new Star, Comet compared April 17 in the Morning 19 322 25 3 20 41 56 7 A new Star, Comet compared April 17 in the Morning 49 323 25 38 17 12 5 3 Delta of Capricorn, Comet estimated April 14 and 15 in the Morning 20 325 10 5 11 26 3 7 A new Star, Comet compared April 7 and 8 in the Morning 21 326 12 50 9 42 0 7 A new Star, Comet compared April 6 in the Morning 30 327 41 29 7 40 52 6 A Star of Aquarius, Comet compared April 1 in the Morning 22 329 11 24 8 41 31 7 A new Star, Comet compared April 2 in the Morning 5 344 4 32 0 48 0 N 6 A of Pisces, Comet compared and estimated February 14 23 348 33 20 1 7 28 8 A new Star, Comet compared January 31 and February 1 8 348 38 36 0 3 40 S 5 Kappa' of Pisces, Comet compared February 3 and 4 24 348 49 10 0 32 29 N 8 A new Star, Comet compared February 1 25 349 40 54 1 1 14 8 A new Star, Comet compared January 27 and 28 26 350 44 16 0 50 56 10 A new Star, Comet compared January 25 27 350 52 16 1 5 8 10 A new Star, Comet compared January 23 16 351 0 1 0 44 48 6 A Star of Pisces, Comet estimated January 25, 27, and 28 28 352 13 5 1 6 40 8 A new Star, Comet compared January 21 and 22 18 352 26 6 0 25 53 5 Lambda of Pisces, Comet estimated January 21
	fwedu_simp:1. 149 48 55 6 7 45 S A new Star was seen on May 18 and 19 at night.
2. 153 25 27 6 52 25 6 A Star of the Sextant was seen on May 14, 15, 16, and 17.
3. 153 33 8 5 12 52 8 A new Star was seen on May 17, 18, 19, and 20.
4. 153 35 49 3 10 3 6 A Star of the Sextant was seen on June 3 and 4.
5. 153 42 34 5 21 5 10 A new Star was seen on May 24, 25, 26, 27, and 28.
6. 153 46 21 7 58 55 8 A new Star was seen on May 14.
7. 153 47 22 7 12 32 10 A new Star was seen on May 14.
8. 153 57 40 8 19 29 10 A new Star was seen on May 13.
9. 154 7 39 4 13 20 10 A new Star was seen on May 30.
10. 154 47 30 12 21 11 7 A new Star was seen on May 7.
11. 154 52 21 11 52 46 9 A new Star was seen on May 8.
12. 154 57 2 13 39 6 7 A new Star was seen on May 6.
13. 156 8 56 15 5 53 6 A Star of the Hydra Phi2 was seen on May 5.
14. 156 8 15 11 1 27 7 A new Star was seen on May 9.
15. 156 43 56 15 37 38 5 Phi3 of the Hydra was seen on May 5.
16. 156 58 8 19 36 31 9 A new Star was seen on May 13.
17. 157 40 41 9 21 39 8 A new Star was seen on May 3.
18. 159 25 32 8 34 41 6 A new Star was seen on May 12.
19. 159 26 5 25 31 55 7 A new Star was seen on May 1.
20. 160 0 28 25 36 24 9 A new Star was seen.
21. 322 3 15 20 52 19 8 A new Star was seen on April 17 in the Morning.
22. 322 25 38 17 12 5 3 Delta of Capricorn was seen on April 14 and 15 in the Morning.
23. 325 10 5 11 26 3 7 A new Star was seen on April 7 and 8 in the Morning.
24. 326 12 50 9 42 0 7 A new Star was seen on April 6 in the Morning.
25. 327 41 29 7 40 52 6 A Star of Aquarius was seen on April 1 in the Morning.
26. 329 11 24 8 41 31 7 A new Star was seen on April 2 in the Morning.
27. 344 4 32 0 48 0 N A of Pisces was seen and estimated on February 14.
28. 348 33 20 1 7 28 8 A new Star was seen on January 31 and February 1.
29. 348 38 36 0 3 40 S Kappa' of Pisces was seen on February 3 and 4.
30. 348 49 10 0 32 29 N A new Star was seen on February 1.
31. 349 40 54 1 1 14 8 A new Star was seen on January 27 and 28.
32. 350 44 16 0 50 56 10 A new Star was seen on January 25.
33. 350 52 16 1 5 8 10 A new Star was seen on January 23.
34. 351 0 1 0 44 48 6 A Star of Pisces was seen and estimated on January 25, 27, and 28.
35. 352 13 5 1 6 40 8 A new Star was seen on January 21.
36. 352 26 6 0 25 53 5 Lambda of Pisces was seen and estimated on January 21.
\end{lstlisting}

\end{document}